\documentclass[10pt,twocolumn,letterpaper]{article}

\usepackage[pagenumbers]{cvpr}
\usepackage[utf8]{inputenc}
\usepackage[T1]{fontenc}
\usepackage{microtype}
\usepackage{graphicx}
\usepackage{amsmath}
\usepackage{amsfonts}
\usepackage{amssymb}
\usepackage{booktabs}
\usepackage{multirow}
\usepackage{caption}
\usepackage{placeins}
\usepackage{subcaption}
\usepackage{algorithm}
\usepackage{algorithmic}
\usepackage{newfloat}
\usepackage{listings}
\usepackage{fancyhdr}
\usepackage{fontawesome5}
\usepackage{xcolor}

\DeclareCaptionStyle{ruled}{
  labelfont=normalfont,
  labelsep=colon,
  strut=off
}
\floatstyle{ruled}
\newfloat{listing}{tb}{lst}{}
\floatname{listing}{Listing}

\definecolor{cvprblue}{rgb}{0.21,0.49,0.74}
\usepackage[pagebackref,breaklinks,colorlinks,allcolors=cvprblue]{hyperref}

\newlength{\yuvionheadextra}
\fancypagestyle{yuvionheader}{%
  \fancyhf{}%
  \fancyhead[L]{\raisebox{0pt}{\includegraphics[height=17pt]{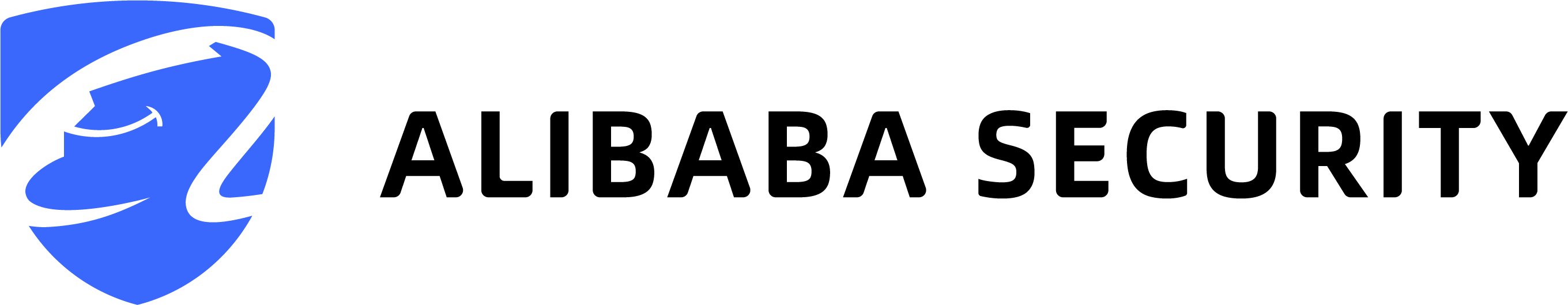}}}%
  \fancyhead[R]{\raisebox{0pt}{\includegraphics[height=17pt]{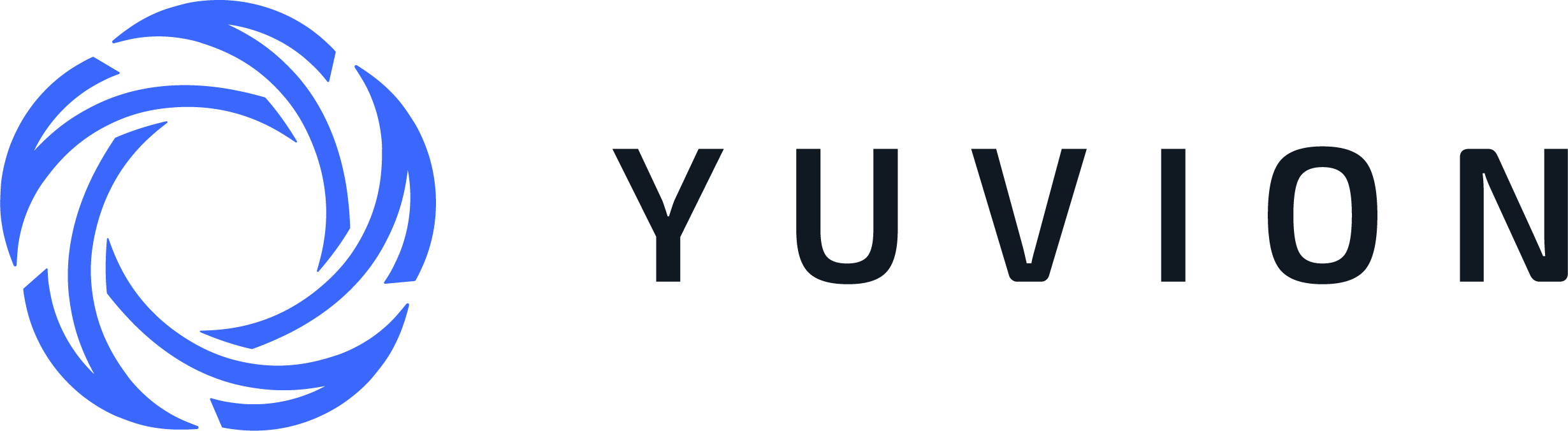}}}%
  \fancyfoot[C]{\thepage}%
}

\def\paperID{}
\def\confName{Yuvion Technical Report}
\def\confYear{2026}
\title{MetaVideoAgent: Automated Video-Agent Evolution for Long-Form Video Understanding}

\author{%
\textbf{Benlei Cui$^{1,*}$, Ruize Wang$^{2,*}$, Junjie Li$^{1,\dagger}$, Jinhao Chen$^{3}$, Yinghao Chen$^{1}$, Longtao Huang$^{1}$}\\[0.10em]
\textbf{Yuwen Zhai$^{1}$, Jingqun Tang$^{4}$, Ruijian Jia$^{1}$, Weiwei Wu$^{2}$, Pengfei Sun$^{1,\ddagger}$, Haiwen Hong$^{1,\dagger}$}\\[0.45em]
\small $^{1}$Yuvion Team, Alibaba Group \quad $^{2}$Zhejiang University\\
\small $^{3}$Beihang University \quad $^{4}$ByteDance\\[0.25em]
\small \href{mailto:cuibenlei.cbl@alibaba-inc.com}{\texttt{cuibenlei.cbl@alibaba-inc.com}} \quad
\href{mailto:honghaiwen.hhw@alibaba-inc.com}{\texttt{honghaiwen.hhw@alibaba-inc.com}}\\[0.2em]
\small \faGithub\;\href{https://github.com/Alibaba-VELLDEPTH/MetaVideoAgent}{\texttt{Alibaba-VELLDEPTH/MetaVideoAgent}}
}

\begin{document}
\raggedbottom
\pagestyle{yuvionheader}
\maketitle
\begingroup
\renewcommand{\thefootnote}{\fnsymbol{footnote}}
\footnotetext[1]{Equal contribution.}
\footnotetext[2]{Corresponding author.}
\footnotetext[3]{Project leader.}
\endgroup
\setcounter{footnote}{0}
\thispagestyle{yuvionheader}

\begin{abstract}
%
%

Long-form video understanding requires locating sparse, question-relevant
evidence in long, multimodal videos. Yet real-world video distributions differ
substantially in modality-specific information density, content structure, and
evidence patterns, causing fixed video-agent designs to incur redundant
processing or even fail when mismatched. Although automated agent evolution has
been explored for text tasks, extending it to video poses three challenges:
full long-video execution makes candidate validation expensive and evolution
inefficient; diverse failures can propagate across coupled evidence-processing
stages, obscuring their root causes; and complex, coupled video preprocessing,
modality-specific perception tools, and evidence-localization strategies make
code-level updates difficult to implement reliably.

We introduce \textbf{MetaVideoAgent}, a framework that automatically evolves a video
agent for a target distribution. To improve evolution efficiency, it first
profiles modality-specific information density and evidence requirements from
sparsely sampled frames and associated queries, providing a low-cost direction
for the initial agent design. During subsequent iterations, it compresses
localized failures from full video-question-answering execution into independently executable minimal
validation tasks for rapidly testing candidate modifications. To diagnose
failures, it constructs a Gold Path using ground-truth answers and annotated
evidence intervals, audits the Student trajectory against this reference, and
aggregates failures across samples to identify recurring causes and their
primary responsible modules. To support reliable code-level updates,
MetaVideoAgent represents a video agent as a modular system and restricts each
iteration to the primary responsible module and its necessary dependencies.

We further introduce \textbf{VA-EvoBench}, covering eight distinct video distributions,
each with relatively concentrated content characteristics and evidence
requirements to support diagnostic analysis of evolution behavior. Each
distribution has separate evolution and held-out splits, and methods
are evaluated on task accuracy, per-question resource use, and evolution efficiency.
With four evolution iterations conducted independently for each distribution,
MetaVideoAgent improves every corresponding initial agent and raises
macro-average accuracy from \textbf{38.44\% to 51.47\%}, at an average
evolution cost of \textbf{3.54M tokens per distribution}. The evolved
distribution-specific agents outperform the best-performing prior fixed-design
video agent by \textbf{6.39 percentage points}, while using the fewest tokens
and video frames per question among the compared video agents. We will release all
code and data to facilitate reproducible research and future work on automated
video-agent evolution.
\end{abstract}

\section{Introduction}

Long-form video understanding is a fundamental problem in multimodal
intelligence. Its central challenge is localizing and correctly perceiving
sparse, question-relevant evidence from extended temporal
contexts~\citep{wu2024longvideobench}. Real-world distributions differ
substantially: e-commerce livestreams convey key information through speech,
sports broadcasts depend on brief visual events, and films require character
and event relations across distant scenes~\citep{chen2024cgbench}.
Consequently, different distributions require different evidence-localization
and perception strategies.

\begin{figure}[t]
    \centering
    \includegraphics[width=0.94\columnwidth]{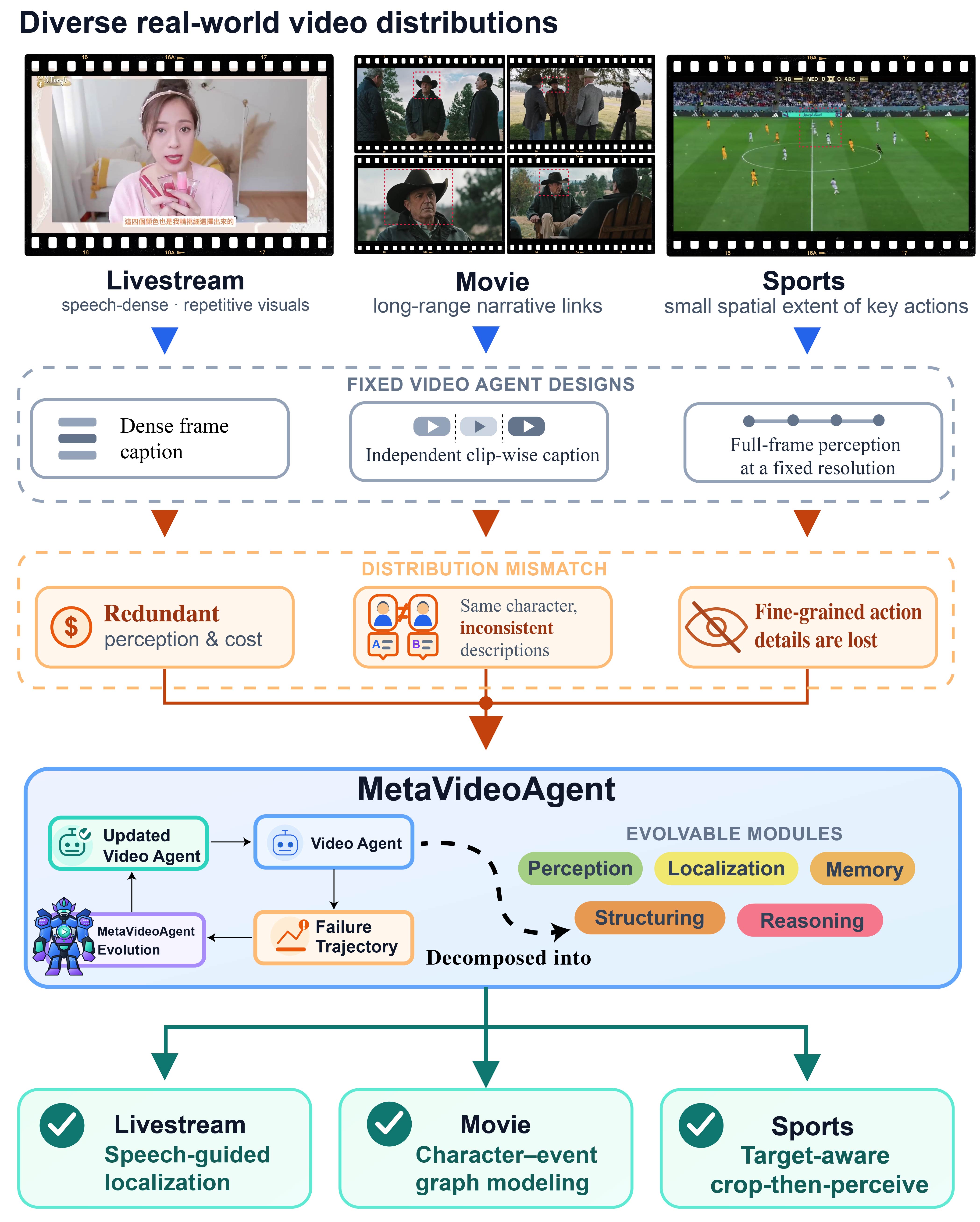}
    \caption{Motivation and overview of MetaVideoAgent. Fixed video-agent
    designs can mismatch diverse video distributions; MetaVideoAgent instead
    automatically evolves a distribution-specific agent design to match the
    dominant evidence patterns of each target distribution.}
    \label{fig:intro_overview}
\end{figure}

Recent video agents use LLMs to iteratively invoke localization and perception
tools for evidence collection~\citep{wang2024videoagent,liu2026videomind}.
However, most employ fixed designs across video distributions. As shown in
Fig.~\ref{fig:intro_overview}, a design suited to one evidence pattern may
incur substantial redundant processing or fail to retrieve answer-critical
evidence under another distribution. This motivates video agents that
automatically evolve their designs for a target video distribution.

%
%
Recent work has explored automated agent evolution for text-based tasks such as coding and language reasoning~\citep{hu2025automated,zhang2025aflow}, whose basic workflow can be summarized as execution, diagnosis, and evolution. Video agents, however, are substantially more complex than text-based agents. Beyond textual information, videos contain visual, auditory, and temporal information, whose channels differ in representation, information density, and temporal granularity. While text agents typically receive textual evidence that can be directly used for reasoning, video agents must actively localize and perceive relevant evidence from long, multimodal video streams. Their evolvable components therefore extend beyond the reasoning process to multiple interdependent stages of evidence processing.

This complexity amplifies the difficulty of all three stages of automated evolution. During execution and validation, repeatedly evaluating every candidate through the complete long-video question-answering pipeline makes the overall evolution process inefficient. During diagnosis, failures may arise from inaccurate audiovisual perception, improper temporal evidence processing, inappropriate context management, or deficient reasoning strategies. These errors may further propagate through the processing pipeline, making their root causes difficult to identify. During evolution, video agents typically involve preprocessing for multiple modalities, complex perception-tool invocation, and evidence-localization strategies. Even with state-of-the-art coding agents such as Claude Code~\citep{anthropic2025claudecode} and OpenAI Codex~\citep{openai2025codex}, reliably modifying such a large, tightly coupled system and validating its end-to-end benefits remains challenging.

To address these challenges, we introduce \textbf{MetaVideoAgent}, an
automated video-agent evolution framework tailored to a target video
distribution. For evolution efficiency, Distribution-Aware Design profiles
sparsely sampled frames and associated queries to guide the initial agent
design, while minimal validation tasks convert localized failures into
independently executable, low-cost validation tasks. For failure diagnosis, the Teacher
Video Agent constructs an evidence-grounded Gold Path and audits the Student
trajectory; the Diagnosis Agent aggregates recurring failures across samples
and identifies their causes and primary responsible modules. For code
evolution, the Evolution Agent performs responsibility-constrained updates over
a modular video-agent representation, modifying dependent modules only when
necessary to preserve interfaces and end-to-end execution.

Beyond the framework, we establish VA-EvoBench, spanning eight substantially
different video distributions with separate evolution and held-out splits.
Each distribution is internally coherent in content structure, modality
density, and evidence requirements, while the distributions remain distinct.
Compared with mixed-type data, this design exposes recurring failure patterns
and provides clearer diagnostic signals for distribution-level design
deficiencies. For each target distribution, a method evolves an agent under a
fixed evolution budget using only the evolution split. Held-out outcomes for
each archived checkpoint are independently recorded and, only after the budget
is exhausted, analyzed for post-hoc evaluation; they are never used for
diagnosis, candidate selection, early stopping, or rollback.
VA-EvoBench measures task accuracy, per-question resource use, and evolution
efficiency under cumulative evolution compute.

%
%
Our main contributions are summarized as follows:
\begin{itemize}
    %
    %
    \item To the best of our knowledge, we are the first to study
    target-distribution-specific automated video-agent evolution for
    long-form video understanding, enabling video agents to adaptively evolve
    their designs for different video distributions. This work promotes a shift
    from manually engineered, heuristic-driven video-agent designs toward an
    adaptive evolution paradigm.

    %
    %
    \item We introduce MetaVideoAgent, an automated video-agent evolution
    framework tailored to a target video distribution. Using the evolution
    split, it efficiently and iteratively improves the agent design for the
    target distribution.

    %
    \item We establish VA-EvoBench, an automated video-agent evolution benchmark spanning
    eight substantially different video distributions. With separate evolution
    and held-out splits for each distribution, VA-EvoBench
    evaluates task accuracy, per-question resource use, and evolution efficiency under
    fixed evolution budgets, providing a unified and reproducible basis for
    research on automated video-agent evolution.
\end{itemize}

\section{Related Work}

\paragraph{Long-Form Video Understanding.}
Multimodal intelligence has advanced rapidly~\citep{cui2026simplepostersimplebaselineproduct,cui2026tcpadetrajectoryconsistentpadeapproximation,cui2026diffusionprobegeneratedimage,qiu2026yuvionvlmultimodalfoundation,liu2025erasediffusion,sun2025attentiveeraserunleashingdiffusion},
while long-form video understanding remains one of its core challenges.
Long-form video benchmarks evaluate models across diverse durations and tasks,
highlighting the difficulty of retrieving sparse, relevant evidence from
extended temporal contexts~\citep{fu2024videomme,wu2024longvideobench,zhou2024mlvu}.
Recent video agents use LLMs to iteratively invoke localization, perception, and
memory tools for evidence collection~\citep{wang2024videoagent,zhang2025deepvideodiscovery,lin2026videoseek}.
Representative systems employ structured memories, multi-granular clip search,
specialized reasoning roles, or multi-agent decomposition~\citep{yeo2025worldmm,wang2025videotree,zhang2024omagent,yan2026symphony}.
Although their actions can adapt to individual queries, the video
representation, tool set, perception granularity, memory organization, and
localization workflow are typically specified manually, limiting adaptation to
distribution-specific evidence-processing requirements.

\paragraph{Adaptive Video Agents.}
Several video agents dynamically adjust temporal scope, sampling density,
memory access, or tool-use policies according to the current query and
evidence~\citep{li2026lenswalk,lin2026videoseek,yeo2025worldmm}.
These methods provide query- or instance-level adaptation within a manually
specified agent design, but do not redesign the full agent pipeline for a target
video distribution.

\paragraph{Automated Design of Agentic Systems.}
Text-domain research has explored automated agent improvement through execution
feedback. ADAS iteratively generates and evaluates agent code, AFlow searches
over LLM-calling workflows, and EvoAgent applies mutation, crossover, and
selection to construct multi-agent systems~\citep{hu2025automated,zhang2025aflow,yuan2025evoagent}.
These methods demonstrate that agent designs can be optimized automatically, but
primarily target text-based reasoning and do not address the multimodal
perception, temporal evidence localization, and costly validation involved in
video agents. MetaVideoAgent extends automated agent design to code-level
video-agent evolution for a target video distribution.

\section{Preliminaries}
\label{sec:preliminaries}

\subsection{Video-Agent Evolution Setup}
\label{sec:evolution_setup}


\paragraph{Evolution data and agent contract.}
Each evolution sample contains a video \(V_i\), a multiple-choice query
\(q_i\) (including its question text and answer options), its ground-truth
answer \(y_i\), and annotated evidence intervals
\(I_i^{\rm ev}\). We denote the evolution set by
\[
\mathcal{D}_{\rm tr}
= \{(V_i,q_i,y_i,I_i^{\rm ev})\}_{i=1}^{N_{\rm tr}}.
\]
An executable video agent \(A\in\mathcal{A}_{\rm exec}\) observes only
\((V_i,q_i)\) and returns an answer \(\hat y_i\), an execution trajectory
\(\tau_i\), and a cost record \(c_i\).


\paragraph{Evolution objective.}
Following ADAS~\citep{hu2025automated}, an evolution algorithm receives an
optional initial agent \(A_{\rm in}\) and the evolution set:
\[
\mathcal{E}(A_{\rm in},\mathcal{D}_{\rm tr})
  = (A^\star,\mathcal{H}),
\qquad
A^\star\in\mathcal{A}_{\rm exec}.
\]
Here, \(\mathcal{H}\) records only the artifacts produced during evolution on
\(\mathcal{D}_{\rm tr}\). Evolution may modify any part of the
code-representable agent design space. After evolution, the resulting agent is
evaluated on a held-out evaluation set \(\mathcal{D}_{\rm te}\) that is disjoint from the
evolution set, i.e., \(\mathcal{D}_{\rm tr}\cap\mathcal{D}_{\rm te}=\emptyset\).

\subsection{Modular Video-Agent Representation}
\label{sec:modular_agent}

Table~\ref{tab:prior-video-agent-module-mapping} in
Appendix~\ref{app:design-space} shows how representative existing
video-agent designs instantiate the proposed framework.
From this review, we observe that most can be represented by five
modules with stable interfaces, denoted by \(A=(S,L,P,W,R)\): video
structuring, evidence localization, perception, working memory, and reasoning.
This modular representation provides a stable basis for video-agent evolution.
Concretely, structuring converts a video into a time-indexed representation,
localization returns executable temporal ranges from this representation,
perception obtains observations only from the returned ranges, working memory
preserves the resulting context across steps, and reasoning plans tool calls
and produces the final answer.  Perception tools provide observations rather
than answer decisions; the unified reasoning module remains responsible for
integrating them and answering the query.
\section{MetaVideoAgent}
\label{sec:metavideoagent}

\subsection{Overview}
\label{sec:method_overview}

\begin{figure*}[t]
    \centering
    \includegraphics[width=\textwidth]{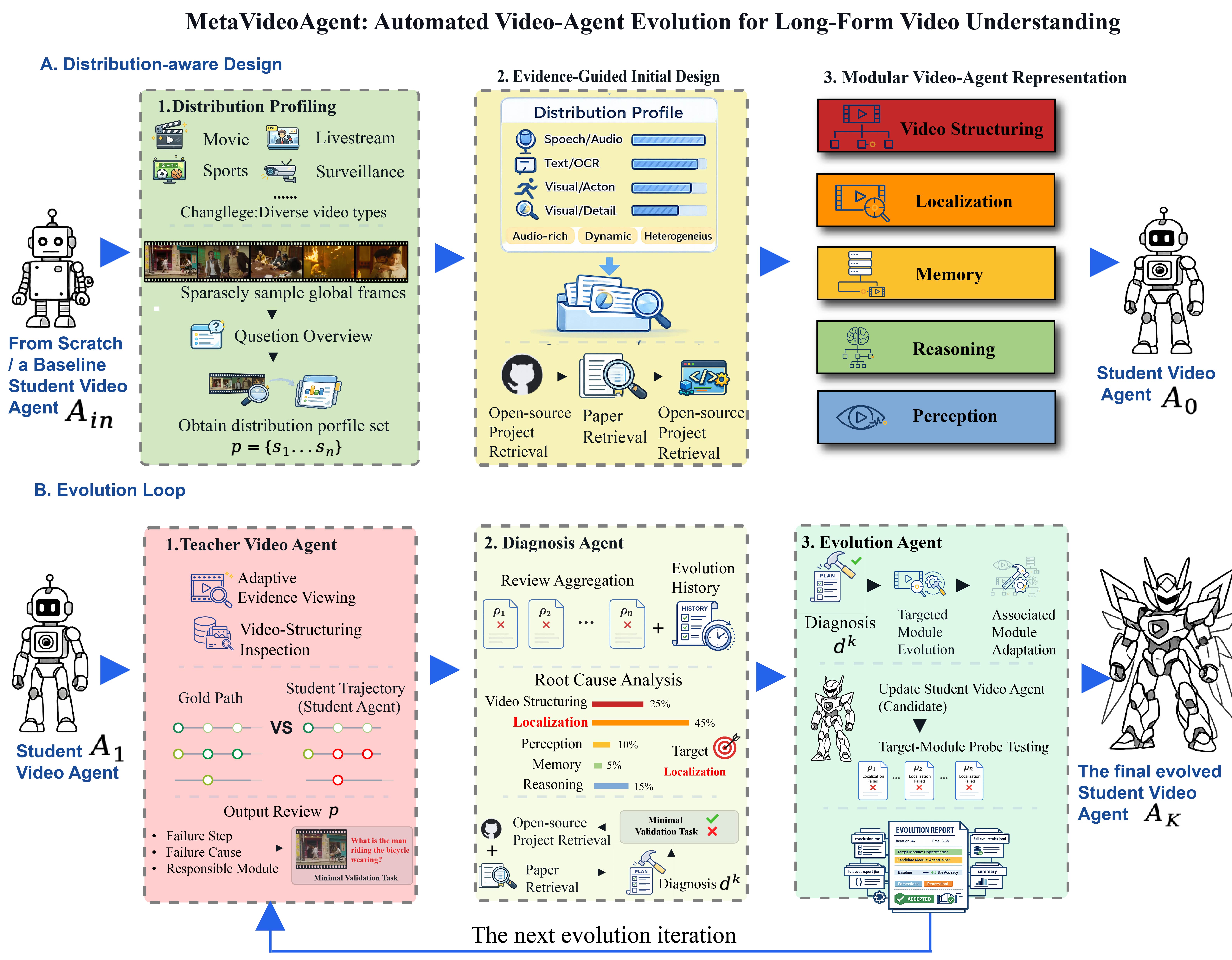}
    \caption{Overview of MetaVideoAgent. Distribution-Aware Design initializes
    the Student Video Agent for a target video distribution. In each iteration,
    the Teacher Video Agent reviews failures, the Diagnosis Agent identifies
    recurring causes and responsible modules, and the Evolution Agent implements
    targeted code-level updates. A validated candidate becomes the Student for
    the next iteration.}
    \label{fig:metavideoagent_overview}
\end{figure*}

MetaVideoAgent first constructs or adapts an initial agent for the target video
distribution. It then iterates through three specialized agents: the
\textbf{Teacher Video Agent} reviews failed Student trajectories, the
\textbf{Diagnosis Agent} identifies recurring causal failures across samples,
and the \textbf{Evolution Agent} implements code-level updates. Each
candidate undergoes engineering checks, minimal validation, and full
evolution-split comparison before promotion. Held-out outcomes for archived
checkpoints are recorded only by an external evaluator for post-hoc reporting;
they never enter diagnosis, promotion, stopping, or rollback.
Figure~\ref{fig:metavideoagent_overview} illustrates the
framework, and Appendix Algorithm~\ref{alg:metavideoagent-full} gives the
complete control flow.

\subsection{Distribution-Aware Design}
\label{sec:initialization}

\paragraph{Distribution profiling.}
For each evolution video \(V_i\), we uniformly sample five frames and
summarize them with non-answer metadata and associated queries
\(\mathcal{Q}_i\) into a record \(s_i\). The target-distribution
profile is \(p=\{s_i\}_{i=1}^{N_v}\). Each record retains overview frames
together with its associated queries and answer options. By jointly surveying
these records, Distribution-Aware Design may abstract recurring
distribution-level patterns relevant to agent design; it does not prescribe a
fixed set of query statistics or modality requirements. Answers and evidence
annotations are not accessed.

\paragraph{Evidence-guided initial design.}
MetaVideoAgent induces distribution-specific requirements \(h_0\) for
perception modalities, evidence granularity, and localization patterns, then
retrieves source-grounded design evidence \(r_0\) from papers, project
pages, and official documentation. The Evolution Agent constructs
\(A_0\) either from an empty implementation or by adapting an existing
video agent such as DVD. The resulting executable agent is repaired and
smoke-tested before evolution. Appendix
Algorithm~\ref{alg:distribution-aware-design} specifies this initialization
procedure.

\subsection{Teacher Video Agent: Question-Level Review}
\label{sec:diagnosis}

At iteration \(k\), the current agent \(A_k\) is evaluated on the
evolution split. For each incorrect sample \(i\), the Teacher receives
\[
x_{k,i}=(\tau_{k,i},q_i,y_i,I_i^{\rm ev},A_k),
\]
where \(\tau_{k,i}\) is the Student trajectory and
\(I_i^{\rm ev}\) is the annotated evidence interval. With access to the
answer and evidence interval, the Teacher constructs an evidence-grounded
\emph{Gold Path} describing the observations and reasoning required for the
correct solution. It then audits the Student trajectory to identify the earliest
step that can no longer support that solution.

The Teacher can revisit raw evidence with adaptive sampling or inspect the
Student's time-aligned video-structuring store. It outputs a review
\(\rho_{k,i}\) containing the earliest unsupported step, causal failure,
responsible module, and a candidate minimal validation task. Detailed tool
interfaces are provided in Appendix~\ref{app:review-validation-details};
the complete review procedure appears in
Algorithm~\ref{alg:teacher-review}.

\paragraph{Minimal validation task.}
The Teacher attempts to isolate each failure as
\(u_{k,i}=(z_{k,i},y_{k,i}^{\rm exp})\), where \(z_{k,i}\) is a
localized multimodal input and \(y_{k,i}^{\rm exp}\) is the expected
output. If reliable isolation is impossible, the original video-QA sample is
retained. These tasks provide low-cost local evidence, while final promotion
still requires full video-QA evaluation. Appendix
Algorithm~\ref{alg:minimal-validation-task} describes their construction and
consolidation.

\subsection{Diagnosis Agent: Cross-Trajectory Diagnosis}
\label{sec:diagnosis_agent}

The Diagnosis Agent aggregates the reviews
\(\mathcal{R}_k=\{\rho_{k,i}\}_{i\in F_k}\), their minimal
tasks, the current design \(A_k\), and evolution history
\(\mathcal{H}_k\). It clusters recurring causal failures, separates
repairable engineering faults from design deficiencies, and selects one primary
responsible module. Previous corrections and regressions are considered to
avoid repeating ineffective directions.

The structured diagnosis is
\[
d_k=(m_k,\phi_k,Q_k,\Gamma_k,\Pi_k),
\]
where \(m_k\) is the primary module, \(\phi_k\) is the recurring
failure, \(Q_k\) is the consolidated minimal validation set,
\(\Gamma_k\) constrains code scope and dependent interfaces, and
\(\Pi_k\) specifies validation and cost requirements. Trajectory
inspection, evidence reinspection, targeted literature retrieval, and
capability validation are invoked only when the aggregated reviews are
insufficient. Further details are given in
Appendix~\ref{app:review-validation-details}, with the aggregation procedure
specified in Algorithm~\ref{alg:cross-trajectory-diagnosis}.

\subsection{Evolution Agent: Modular Code Modification}
\label{sec:minimal_validation}

The Evolution Agent implements \(d_k\) by modifying the primary module
\(m_k\). Dependent modules are changed only when required to preserve
interfaces or end-to-end execution, producing a complete candidate
\(\widetilde A_k\).

\paragraph{Candidate acceptance.}
Let \(s_k=s(A_k,\mathcal{D}_{\rm tr})\) and
\(\widetilde s_k=s(\widetilde A_k,\mathcal{D}_{\rm tr})\)
denote task, cost, and validity evidence. The decision is
\[
g_k=G_k^{\rm eng}\land G_k^{\rm probe}
\land\mathrm{Accept}(\widetilde s_k,s_k).
\]
Engineering and minimal-validation gates may reject a candidate early, but only
full evolution-split comparison can approve promotion. If \(g_k=1\),
\(A_{k+1}=\widetilde A_k\); otherwise \(A_{k+1}=A_k\). Reviews,
diagnoses, code diffs, validation traces, and decisions are retained in
\(\mathcal{H}_{k+1}\). The accepted checkpoint is retained for external
held-out reporting only; the result cannot affect any evolution decision. The
complete responsibility-constrained update, repair,
probe, and promotion procedure appears in Appendix
Algorithm~\ref{alg:responsibility-constrained-evolution}.

\section{Video-Agent Evolution Benchmark}
\label{sec:benchmark}

We introduce VA-EvoBench, derived from CG-Bench~\citep{chen2024cgbench}, with
eight target distributions and video-disjoint evolution and held-out splits.
Under a fixed budget, each method evolves one agent per distribution. The final
agent $A_K$ determines primary held-out accuracy; an external evaluator records
intermediate held-out checkpoints only for post-hoc analysis, never exposing
them to evolution. We report held-out accuracy, per-question deployment cost,
and evolution efficiency.

\begin{table*}[!t]
\centering
\scriptsize
\setlength{\tabcolsep}{2.6pt}
\renewcommand{\arraystretch}{0.8}
\resizebox{\textwidth}{!}{%
\begin{tabular}{lccccccccccc}
\toprule
Method & Int. & Prod. & Soft. & Drama & Stage & Game & Sport & Course
& Avg. & Tok./q (K) $\downarrow$ & Frm./q $\downarrow$ \\
\midrule
\multicolumn{12}{c}{\textit{Closed-source MLLMs}} \\
\midrule
GPT-5.6-Sol & 21.48 & 23.30 & 30.30 & 16.81 & 31.07 & 21.18 & 15.49 & 36.97 & 24.58 & 8.37 & 50 \\
Claude Opus 4.8 & 22.22 & 29.13 & 31.82 & 21.85 & 26.21 & 16.47 & 26.76 & 37.82 & 26.53 & 16.32 & 100 \\
GPT-5.5 & 22.96 & 32.04 & 38.64 & 18.49 & 28.16 & 23.53 & 18.31 & 41.18 & 27.91 & 8.37 & 50 \\
Gemini 3.1 Pro & 36.30 & 41.75 & 43.94 & 22.69 & 45.63 & 30.59 & 33.80 & \underline{57.14} & 38.98 & 19.50 & 120 \\
\midrule
\multicolumn{12}{c}{\textit{Open-source MLLMs}} \\
\midrule
Qwen3-VL-235B-Instruct & 22.96 & 27.18 & 28.03 & 16.81 & 25.24 & 15.29 & 23.94 & 31.93 & 23.92 & 19.50 & 120 \\
LLaVA-Video-72B & 24.44 & 26.21 & 32.58 & 17.65 & 24.27 & 20.00 & 22.54 & 30.25 & 24.74 & 19.50 & 120 \\
Qwen3-VL-235B-Thinking & 28.89 & 29.13 & 40.15 & 23.53 & 39.81 & 23.53 & 29.58 & 37.82 & 31.55 & 19.50 & 120 \\
\midrule
\multicolumn{12}{c}{\textit{Fixed Video Agents}} \\
\midrule
M3-Agent~\citep{long2026m3agent} & 31.11 & 33.98 & 35.61 & 25.21 & 25.24 & 28.23 & 26.76 & 41.18 & 30.92 & 131.94 & 36.23 \\
Symphony~\citep{yan2026symphony} & \textbf{48.15} & \underline{57.28} & 48.48 & 35.29 & 46.60 & 18.82 & 32.39 & 43.70 & 41.34 & 322.94 & 830.18 \\
DVD~\citep{zhang2025dvd} & 45.19 & \underline{57.28} & 40.91 & \textbf{47.06} & \underline{52.43} & 30.59 & 30.99 & 48.74 & 44.15 & 214.19 & 38.75 \\
WorldMM~\citep{yeo2025worldmm} & 40.00 & \textbf{59.22} & \underline{52.27} & \underline{45.38} & 48.54 & \underline{31.76} & \underline{42.25} & 41.18 & \underline{45.08} & 102.26 & 42.77 \\
\midrule
\multicolumn{12}{c}{\textit{Automatically Evolved Video Agents (Ours)}} \\
\midrule
\textbf{MetaVideoAgent} & \underline{47.41} & 56.31 & \textbf{53.03} & 40.34 & \textbf{57.28} & \textbf{47.06} & \textbf{50.70} & \textbf{59.66} & \textbf{51.47} & 74.47 & 29.11 \\
\bottomrule
\end{tabular}%
}
\renewcommand{\arraystretch}{1.0}
\caption{Main results across video distributions. Distribution columns report
QA accuracy (\%). Avg. is the macro average over the eight target distributions.
Tok./q and Frm./q report the per-question token and image-frame statistics.
Bold and underlined entries denote the best and second-best accuracy values,
respectively.
For MVA, each distribution accuracy reports the agent obtained at the end of evolution.}
\label{tab:main_results}
\end{table*}

\subsection{Target Distributions}

Videos are grouped by recurring content form, modality composition, and
evidence-localization requirements so that failures can be aggregated into
distribution-level design deficiencies. The eight distributions span dialogue-,
object-, screen-, narrative-, action-, and instruction-centric evidence and are
evolved independently. Appendix~\ref{app:target-video-distributions} gives the
construction criteria, released manifests, split statistics, and complete
category definitions.

\subsection{Protocol and Measures}

On the evolution split, the Student observes only the video, question, and
answer options; labels and evidence intervals are released only to the Teacher
after execution. Held-out outcomes never affect review, diagnosis, candidate
selection, stopping, or rollback, and the final agent $A_K$ is reported after a
fixed $K$ iterations. We measure held-out macro accuracy, tokens per question,
and frames per question. With stage-$j$ token cost $c_j$, evolution efficiency
is the best evolution-split accuracy attained under cumulative cost:
\[
C_k=\sum_{j=0}^{k}c_j,\qquad
P_d(C_k)=\max_{0\leq j\leq k}\mathrm{Acc}^{\mathrm{evol}}_d(A_j).
\]
This curve never selects the final agent. Appendices~\ref{app:evolution_protocol}
and~\ref{app:evaluation_measures} provide the complete annotation-access and
accounting rules.

\section{Experiments}
\label{sec:experiments}

We test whether MetaVideoAgent evolves distribution-specific designs, improves
throughout evolution, and depends on each component of the proposed procedure.

\subsection{Experimental Setup}
\label{sec:experimental_setup}

We evolve one agent from scratch per target distribution and compare the final
agents $A_K$ with direct MLLMs and fixed video agents under VA-EvoBench. Appendix~\ref{app:experimental-setup}
provides model assignments, inputs, tools, implementation settings, and cost accounting.

\begin{figure*}[!t]
\centering
\includegraphics[width=\textwidth]{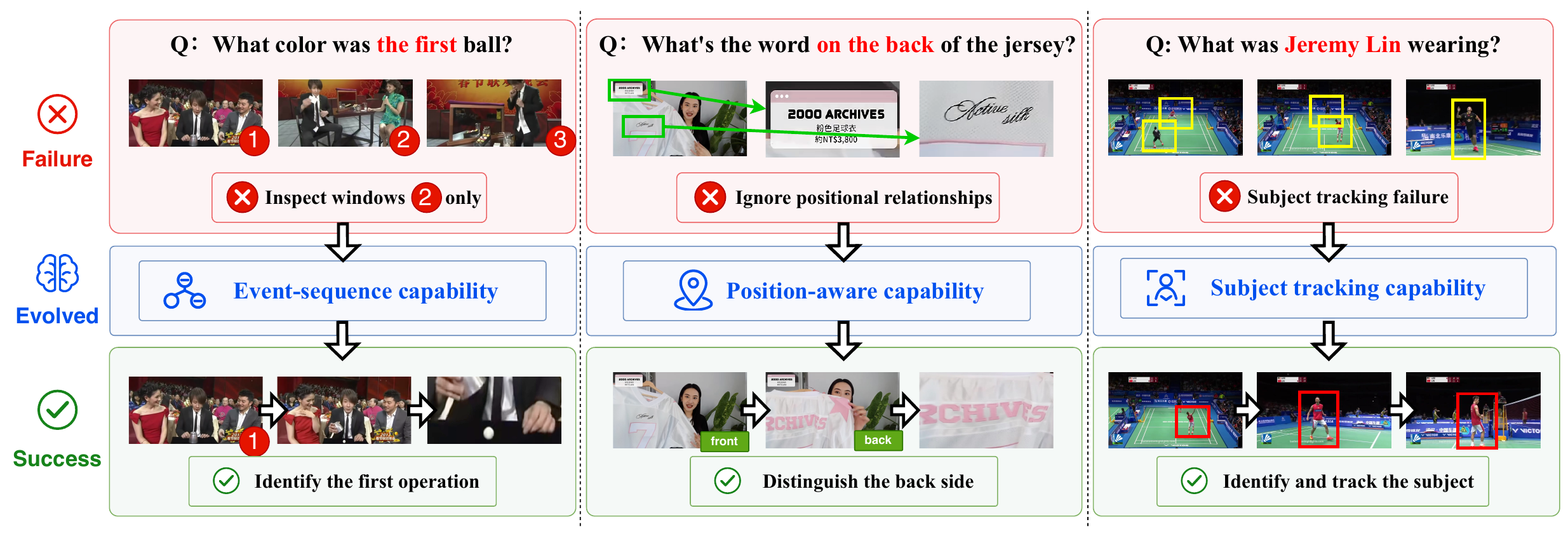}
\caption{Qualitative evolution cases for event ordering, position-relation
perception, and subject tracking. Rows show the initial failure, evolved
capability, and successful held-out application.}
\label{fig:qualitative_cases}
\end{figure*}

\subsection{Main Results}
\label{sec:main_results}

Table~\ref{tab:main_results} shows that MVA reaches 51.47\% macro accuracy
(446/867; 51.44\% question-weighted), 12.49 points above the strongest direct
MLLM, Gemini 3.1 Pro. It improves all eight distributions and exceeds every
fixed video agent while using the fewest per-question tokens and frames.

\subsection{Qualitative Evolution Cases}
\label{sec:qualitative_cases}

Figure~\ref{fig:qualitative_cases} shows three held-out transfers of reusable
behaviors learned without feedback from the displayed questions.

\paragraph{Event ordering: Stage Performance.}
The initial agent treats retrieved candidate windows as unordered and cannot
resolve ``the first ball in the first trick.'' The evolved design orders events
for ordinal queries before perceiving the selected event, identifying the
opening ball as \emph{White}.

\paragraph{Position-relation perception: Product Presentation.}
The evolved design first distinguishes the jersey's front from back and then
reads the requested back-side text as ``RCHIVES.''

\paragraph{Subject tracking: Sports Broadcast.}
Persistent subject identity lets the evolved agent associate Jeremy Lin across
camera views and recognize his attire reliably. These cases cover temporal
order, position relations, and cross-view tracking rather than memorization.

\subsection{Evolution Efficiency}
\label{sec:evolution_dynamics}

\begin{figure}[htbp]
\centering
\includegraphics[width=\columnwidth]{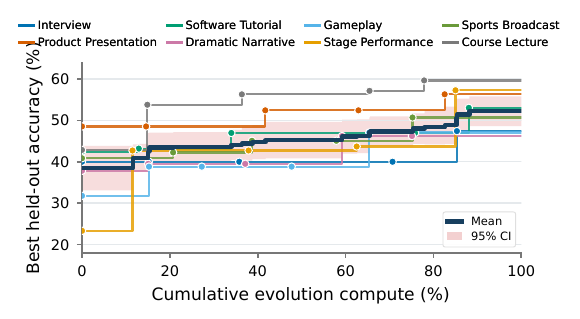}
\caption{Held-out budget--accuracy curves across eight video distributions.
The bold curve and shading denote the mean and 95\% confidence interval.}
\label{fig:evolution_dynamics}
\end{figure}

We plot post-hoc best-so-far held-out accuracy $P_d(C_k)$ against cumulative
evolution cost $C_k$, normalized by each run's total cost
(Figure~\ref{fig:evolution_dynamics}). The curves are constructed only after the
fixed budget is exhausted from externally archived checkpoints; they never
enter evolution or select the final agent $A_K$.

We aggregate right-continuous curves on a common 0--100\% budget grid. Mean
best accuracy at full budget is 52.21\% (95\% CI: 48.63--55.78), whereas final
macro accuracy is 51.47\% because the curve summarizes the best archived
checkpoint. Across eight runs, evolution consumes 28.33M tokens in total
(3.54M per distribution) and raises final macro accuracy from 38.44\% to
51.47\% (+13.03 points).

\subsection{Ablation and Initialization Analysis}
\label{sec:ablations}

We ablate Course Lecture on the same 119-question held-out split
(Table~\ref{tab:ablations}). Removing Diagnosis and targeted research reduces
accuracy by 2.52 and 10.92 points, respectively. The modest short-horizon
Diagnosis gap does not imply dispensability: cross-trajectory aggregation gives
later updates a consistent direction. Review, Diagnosis, and Research are
removed jointly because the latter two consume review outputs. This blind
variant is highly unstable, with 293.24 $\mathrm{pp}^2$ update volatility
versus 15.52 for full MVA. The DVD-initialized variant reaches 63.03\% (+3.37),
showing that MVA can further adapt an existing agent. The probe has no numerical
ablation because it defines both local validation and runtime diagnosis.
Appendix~\ref{app:ablation-trajectories} reports complete trajectories and
further analysis.

\begin{table}[htbp]
\centering
\scriptsize
\setlength{\tabcolsep}{2.4pt}
\renewcommand{\arraystretch}{0.98}
\begin{tabular}{@{}lrrrrr@{}}
\toprule
Setting & Acc. & $\Delta$ & Tok./q (K) & Frm./q & Vol. \\
\midrule
Full MVA & 59.66 & -- & 76.13 & 38.84 & 15.52 \\
w/o Diagnosis & 57.14 & -2.52 & 125.95 & 32.34 & 18.48 \\
w/o Deep Research & 48.74 & -10.92 & 62.02 & 9.61 & 20.59 \\
w/o Review/Diagnosis/Research & 47.90 & -11.76 & 61.09 & 1.90 & 293.24 \\
w/o Min.-Val. Probe & -- & -- & -- & -- & -- \\
MVA from DVD & 63.03 & +3.37 & 75.28 & 2.21 & 131.43 \\
\bottomrule
\end{tabular}
\caption{Ablation and initialization on Course Lecture. $\Delta$ is relative
to full MVA; Tok. and Frm. are per-question costs, and Vol. is variance
($\mathrm{pp}^2$) of consecutive held-out accuracy changes.}
\label{tab:ablations}
\end{table}

\paragraph{Discussion and limitations.}
Appendices~\ref{app:discussion} and~\ref{app:limitations} discuss open
challenges in automated video-agent evolution and outline promising directions
for future research.

\section{Conclusion}

We introduced MetaVideoAgent for target-distribution-specific automated
video-agent evolution and VA-EvoBench for evaluating this process. In our
experiments, four independent evolution iterations across eight target distributions improve
macro-average accuracy from 38.44\% to 51.47\% (+13.03), with gains on every
distribution. The resulting distribution-specific agents outperform the
strongest fixed-design video agent by 6.39 percentage points while achieving the lowest
per-question token and frame costs among the compared video agents. These
results demonstrate that video-agent designs can be evolved from
distribution-level feedback rather than manually fixed in advance.

\clearpage
\bibliographystyle{ieeenat_fullname}
\bibliography{references/aaai2027}

@misc{li2026lenswalk,
  title = {LensWalk: Agentic Video Understanding by Planning How You See in Videos},
  author = {Li, Keliang and Li, Yansong and Shen, Hongze and Liu, Mengdi and Chang, Hong and Shan, Shiguang},
  year = {2026},
  eprint = {2603.24558},
  archivePrefix = {arXiv},
  primaryClass = {cs.CV}
}

@misc{yeo2025worldmm,
  title = {WorldMM: Dynamic Multimodal Memory Agent for Long Video Reasoning},
  author = {Yeo, Woongyeong and Kim, Kangsan and Yoon, Jaehong and Hwang, Sung Ju},
  year = {2025},
  eprint = {2512.02425},
  archivePrefix = {arXiv},
  primaryClass = {cs.CV}
}

@misc{lin2026videoseek,
  title = {VideoSeek: Long-Horizon Video Agent with Tool-Guided Seeking},
  author = {Lin, Jingyang and Wu, Jialian and Liu, Jiang and Sun, Ximeng and Wang, Ze and Yu, Xiaodong and Luo, Jiebo and Liu, Zicheng and Barsoum, Emad},
  year = {2026},
  eprint = {2603.20185},
  archivePrefix = {arXiv},
  primaryClass = {cs.CV}
}

@misc{zhang2025dvd,
  title = {Deep Video Discovery: Agentic Search with Tool Use for Long-form Video Understanding},
  author = {Zhang, Xiaoyi and Jia, Zhaoyang and Guo, Zongyu and Li, Jiahao and Li, Bin and Li, Houqiang and Lu, Yan},
  year = {2025},
  eprint = {2505.18079},
  archivePrefix = {arXiv},
  primaryClass = {cs.CV}
}

@misc{yin2025videoarm,
  title = {VideoARM: Agentic Reasoning over Hierarchical Memory for Long-Form Video Understanding},
  author = {Yin, Yufei and Meng, Qianke and Chen, Minghao and Ding, Jiajun and Shao, Zhenwei and Yu, Zhou},
  year = {2025},
  eprint = {2512.12360},
  archivePrefix = {arXiv},
  primaryClass = {cs.CV}
}

@misc{yin2026haven,
  title = {Hierarchical Long Video Understanding with Audiovisual Entity Cohesion and Agentic Search},
  author = {Yin, Xinlei and Peng, Xiulian and Li, Xiao and Xiong, Zhiwei and Lu, Yan},
  year = {2026},
  eprint = {2601.13719},
  archivePrefix = {arXiv},
  primaryClass = {cs.CV}
}

@misc{fu2024videomme,
  title = {Video-MME: The First-Ever Comprehensive Evaluation Benchmark of Multi-modal LLMs in Video Analysis},
  author = {Fu, Chaoyou and Dai, Yuhan and Luo, Yongdong and Li, Lei and Ren, Shuhuai and Zhang, Renrui and Wang, Zihan and Zhou, Chenyu and Shen, Yunhang and Zhang, Mengdan and Chen, Peixian and Li, Yanwei and Lin, Shaohui and Zhao, Sirui and Li, Ke and Xu, Tong and Zheng, Xiawu and Chen, Enhong and Ji, Rongrong and Sun, Xing},
  year = {2024},
  eprint = {2405.21075},
  archivePrefix = {arXiv},
  primaryClass = {cs.CV}
}

@misc{wang2024videoagent,
  title = {VideoAgent: Long-form Video Understanding with Large Language Model as Agent},
  author = {Wang, Xiaohan and Zhang, Yuhui and Zohar, Orr and Yeung-Levy, Serena},
  year = {2024},
  eprint = {2403.10517},
  archivePrefix = {arXiv},
  primaryClass = {cs.CV},
  doi = {10.48550/arXiv.2403.10517}
}

@misc{wu2024longvideobench,
  title = {LongVideoBench: A Benchmark for Long-context Interleaved Video-Language Understanding},
  author = {Wu, Haoning and Li, Dongxu and Chen, Bei and Li, Junnan},
  year = {2024},
  eprint = {2407.15754},
  archivePrefix = {arXiv},
  primaryClass = {cs.CV}
}

@misc{zhou2024mlvu,
  title = {MLVU: Benchmarking Multi-task Long Video Understanding},
  author = {Zhou, Junjie and Shu, Yan and Zhao, Bo and Wu, Boya and Liang, Zhengyang and Xiao, Shitao and Qin, Minghao and Yang, Xi and Xiong, Yongping and Zhang, Bo and Huang, Tiejun and Liu, Zheng},
  year = {2024},
  eprint = {2406.04264},
  archivePrefix = {arXiv},
  primaryClass = {cs.CV}
}

@inproceedings{zhang2025deepvideodiscovery,  title = {Deep Video Discovery: Agentic Search with Tool Use for Long-form Video Understanding},  author = {Zhang, Xiaoyi and Jia, Zhaoyang and Guo, Zongyu and Li, Jiahao and Li, Bin and Li, Houqiang and Lu, Yan},  booktitle = {Advances in Neural Information Processing Systems},  year = {2025},  eprint = {2505.18079},  archivePrefix = {arXiv},  primaryClass = {cs.CV}}

@inproceedings{liu2026videomind,  title = {{VideoMind}: A Chain-of-{LoRA} Agent for Temporal-Grounded Video Reasoning},  author = {Liu, Ye and Lin, Kevin Qinghong and Chen, Chang Wen and Shou, Mike Zheng},  booktitle = {International Conference on Learning Representations},  year = {2026},  eprint = {2503.13444},  archivePrefix = {arXiv},  primaryClass = {cs.CV}}

@inproceedings{hu2025automated,  title = {Automated Design of Agentic Systems},  author = {Hu, Shengran and Lu, Cong and Clune, Jeff},  booktitle = {International Conference on Learning Representations},  year = {2025},  url = {https://openreview.net/forum?id=t9U3LW7JVX}}

@inproceedings{zhang2025aflow,  title = {{AFlow}: Automating Agentic Workflow Generation},  author = {Zhang, Jiayi and Xiang, Jinyu and Yu, Zhaoyang and Teng, Fengwei and Chen, Xionghui and Chen, Jiaqi and Zhuge, Mingchen and Cheng, Xin and Hong, Sirui and Wang, Jinlin and Zheng, Bingnan and Liu, Bang and Luo, Yuyu and Wu, Chenglin},  booktitle = {International Conference on Learning Representations},  year = {2025},  url = {https://openreview.net/forum?id=z5uVAKwmjf}}

@misc{chen2024cgbench,
  title = {CG-Bench: Clue-grounded Question Answering Benchmark for Long Video Understanding},
  author = {Chen, Guo and Liu, Yicheng and Huang, Yifei and He, Yuping and Pei, Baoqi and Xu, Jilan and Wang, Yali and Lu, Tong and Wang, Limin},
  year = {2024},
  eprint = {2412.12075},
  archivePrefix = {arXiv},
  primaryClass = {cs.CV},
  doi = {10.48550/arXiv.2412.12075}
}

@inproceedings{wang2025videotree,  title = {{VideoTree}: Adaptive Tree-based Video Representation for {LLM} Reasoning on Long Videos},  author = {Wang, Ziyang and Yu, Shoubin and Stengel-Eskin, Elias and Yoon, Jaehong and Cheng, Feng and Bertasius, Gedas and Bansal, Mohit},  booktitle = {Proceedings of the IEEE/CVF Conference on Computer Vision and Pattern Recognition},  pages = {3272--3282},  year = {2025},  doi = {10.1109/CVPR52734.2025.00311},  eprint = {2405.19209},  archivePrefix = {arXiv}}

@inproceedings{chu2025graphvideoagent,  title = {{GraphVideoAgent}: Enhancing Long-form Video Understanding with Entity Relation Graphs},  author = {Chu, Meng and Li, Yicong and Chua, Tat-Seng},  booktitle = {Proceedings of the 33rd ACM International Conference on Multimedia},  pages = {4639--4648},  year = {2025},  doi = {10.1145/3746027.3755537},  eprint = {2501.15953},  archivePrefix = {arXiv}}

@inproceedings{zhang2024omagent,  title = {{OmAgent}: A Multi-modal Agent Framework for Complex Video Understanding with Task Divide-and-Conquer},  author = {Zhang, Lu and Zhao, Tiancheng and Ying, Heting and Ma, Yibo and Lee, Kyusong},  booktitle = {Proceedings of the 2024 Conference on Empirical Methods in Natural Language Processing},  pages = {10031--10045},  year = {2024},  publisher = {Association for Computational Linguistics},  doi = {10.18653/v1/2024.emnlp-main.559},  url = {https://aclanthology.org/2024.emnlp-main.559/}}

@inproceedings{yuan2025evoagent,
  title = {{EvoAgent}: Towards Automatic Multi-Agent Generation via Evolutionary Algorithms},
  author = {Yuan, Siyu and Song, Kaitao and Chen, Jiangjie and Tan, Xu and Li, Dongsheng and Yang, Deqing},
  booktitle = {Proceedings of the 2025 Conference of the Nations of the Americas Chapter of the Association for Computational Linguistics: Human Language Technologies},
  year = {2025},
  eprint = {2406.14228},
  archivePrefix = {arXiv},
  primaryClass = {cs.AI},
  url = {https://arxiv.org/abs/2406.14228}
}

@misc{anthropic2025claudecode,
  author       = {{Anthropic}},
  title        = {Claude Code},
  year         = {2025},
  howpublished = {\url{https://github.com/anthropics/claude-code}},
  note         = {Accessed: 2026-07-29}
}

@misc{openai2025codex,
  author       = {{OpenAI}},
  title        = {OpenAI Codex},
  year         = {2025},
  howpublished = {\url{https://github.com/openai/codex}},
  note         = {Accessed: 2026-07-29}
}

@inproceedings{long2026m3agent,
  title = {Seeing, Listening, Remembering, and Reasoning: A Multimodal Agent with Long-Term Memory},
  author = {Long, Lin and He, Yichen and Ye, Wentao and Pan, Yiyuan and Lin, Yuan and Li, Hang and Zhao, Junbo and Li, Wei},
  booktitle = {International Conference on Learning Representations},
  year = {2026},
  eprint = {2508.09736},
  archivePrefix = {arXiv},
  primaryClass = {cs.CV},
  url = {https://arxiv.org/abs/2508.09736}
}

@inproceedings{yan2026symphony,
  title = {Symphony: A Cognitively-Inspired Multi-Agent System for Long-Video Understanding},
  author = {Yan, Haiyang and Zhou, Hongyun and Xu, Peng and Feng, Xiaoxue and Liu, Mengyi},
  booktitle = {Proceedings of the IEEE/CVF Conference on Computer Vision and Pattern Recognition},
  year = {2026},
  url = {https://openaccess.thecvf.com/content/CVPR2026/html/Yan_Symphony_A_Cognitively-Inspired_Multi-Agent_System_for_Long-Video_Understanding_CVPR_2026_paper.html}
}

@misc{cui2026simplepostersimplebaselineproduct,
  title = {simpleposter: a simple baseline for product poster generation},
  author = {Cui, Benlei and Zeng, Fangao and Jiang, Weitao and Zhai, Yuwen and Hong, Haiwen and Huang, Longtao and Xue, Hui and Shang, Wenxiang and Huang, Pipei},
  year = {2026},
  eprint = {2605.08784},
  archivePrefix = {arXiv},
  primaryClass = {cs.CV},
  url = {https://arxiv.org/abs/2605.08784}
}

@misc{cui2026tcpadetrajectoryconsistentpadeapproximation,
  title = {{TC-Pad'e}: Trajectory-Consistent {Pad'e} Approximation for Diffusion Acceleration},
  author = {Cui, Benlei and He, Shaoxuan and Huang, Bukun and Ye, Zhizeng and Sun, Yunyun and Huang, Longtao and Xue, Hui and Yang, Yang and Tang, Jingqun and Zhao, Zhou and Hong, Haiwen},
  year = {2026},
  eprint = {2603.02943},
  archivePrefix = {arXiv},
  primaryClass = {cs.CV},
  url = {https://arxiv.org/abs/2603.02943}
}

@misc{cui2026diffusionprobegeneratedimage,
  title = {Diffusion Probe: Generated Image Result Prediction Using {CNN} Probes},
  author = {Cui, Benlei and Huang, Bukun and Ye, Zhizeng and Dong, Xuemei and Chen, Tuo and Xue, Hui and Yang, Dingkang and Huang, Longtao and Tang, Jingqun and Hong, Haiwen},
  year = {2026},
  eprint = {2602.23783},
  archivePrefix = {arXiv},
  primaryClass = {cs.CV},
  url = {https://arxiv.org/abs/2602.23783}
}

@misc{qiu2026yuvionvlmultimodalfoundation,
  title = {Yuvion {VL}: A Multimodal Foundation Model for Adversarial Content and {AI} Safety},
  author = {Qiu, Shikai and Xu, Xiaowen and Cui, Benlei and Ma, Ting and Huang, Xiufeng and Jiang, Wenjing and He, Shaoxuan and Xu, Haolei and Chai, Chunyang and Li, Yujian and Zhang, Yiliang and Wang, Guanghui and Wang, Ziheng and Xu, Ziwen and Fan, Zhaoyu and Chen, Jinhao and Jian, Ruijie and Li, Hongxing and Xiao, Chuxi and Chen, Xinyue and Liu, Wenxuan and Dong, Libin and Cao, Yupeng and Xia, Xiaoqian and Wang, Jing and Jiang, Zhe and Ye, Zhenan and Yang, Guang and Liu, Bin and Peng, Wei and Zhu, Ziqiang and Lian, Meihui and Kacuila, Kaiwen Lv and Ding, Haidong and Zhang, Dongjie and Zhou, Yangfan and Zhu, Bingyu and Wang, Yan and Zhao, Hai and Jin, Xuan and Zhao, Wei and Sun, Pengfei and Zhang, Huiming and Wang, Wei and Cao, Xipeng and Chen, Jialun and Chen, Xiao and Ren, Shaola and Hu, Yunqing and Li, Bin and Yao, Chengwen and Huang, Meng and Li, Xianfeng and Tang, Bin and Liu, Chao and Xue, Hui and Huang, Longtao and Hong, Haiwen},
  year = {2026},
  eprint = {2606.25034},
  archivePrefix = {arXiv},
  primaryClass = {cs.CV},
  url = {https://arxiv.org/abs/2606.25034}
}

@inproceedings{liu2025erasediffusion,
  title = {Erase Diffusion: Empowering Object Removal Through Calibrating Diffusion Pathways},
  author = {Liu, Y. and Zhou, H. and Cui, Benlei and Shang, Wenxiang and Lin, R.},
  booktitle = {2025 IEEE/CVF Conference on Computer Vision and Pattern Recognition (CVPR)},
  pages = {2418--2427},
  year = {2025},
  doi = {10.1109/CVPR52734.2025.00231}
}

@misc{sun2025attentiveeraserunleashingdiffusion,
  title = {Attentive Eraser: Unleashing Diffusion Model's Object Removal Potential via Self-Attention Redirection Guidance},
  author = {Sun, Wenhao and Cui, Benlei and Dong, Xue-Mei and Tang, Jingqun},
  year = {2025},
  eprint = {2412.12974},
  archivePrefix = {arXiv},
  primaryClass = {cs.CV},
  url = {https://arxiv.org/abs/2412.12974}
}

@inproceedings{xu2026seeingnotthinking,
  title = {Seeing but Not Thinking: Routing Distraction in Multimodal Mixture-of-Experts},
  author = {Xu, Haolei and Hong, Haiwen and Li, Hongxing and Zhou, Rui and Zhang, Yang and Huang, Longtao and Xue, Hui and Shen, Yongliang and Lu, Weiming and Zhuang, Yueting},
  booktitle = {Proceedings of the 64th Annual Meeting of the Association for Computational Linguistics (Volume 1: Long Papers)},
  pages = {31164--31178},
  year = {2026},
  publisher = {Association for Computational Linguistics},
  doi = {10.18653/v1/2026.acl-long.1438},
  url = {https://doi.org/10.18653/v1/2026.acl-long.1438}
}

@inproceedings{wang2023openvocabularyobjectdetection,
  title = {Open-Vocabulary Object Detection With an Open Corpus},
  author = {Wang, Jiong and Zhang, Huiming and Hong, Haiwen and Jin, Xuan and He, Yuan and Xue, Hui and Zhao, Zhou},
  booktitle = {2023 IEEE/CVF International Conference on Computer Vision (ICCV)},
  pages = {6736--6746},
  year = {2023},
  publisher = {IEEE},
  doi = {10.1109/ICCV51070.2023.00622},
  url = {https://doi.org/10.1109/ICCV51070.2023.00622}
}

@inproceedings{hu2021ramstrans,
  title = {{RAMS-Trans}: Recurrent Attention Multi-scale Transformer for Fine-grained Image Recognition},
  author = {Hu, Yunqing and Jin, Xuan and Zhang, Yin and Hong, Haiwen and Zhang, Jingfeng and He, Yuan and Xue, Hui},
  booktitle = {Proceedings of the 29th ACM International Conference on Multimedia},
  pages = {4239--4248},
  year = {2021},
  publisher = {ACM},
  doi = {10.1145/3474085.3475561},
  url = {https://doi.org/10.1145/3474085.3475561}
}

@misc{li2026perceivetoreason,
  title = {Perceive-to-Reason: Decoupling Perception and Reasoning for Fine-Grained Visual Reasoning},
  author = {Li, Hongxing and Huang, Xiufeng and Li, Dingming and Jiang, Wenjing and Wang, Zixuan and Xu, Haolei and Zhang, Hanrong and Hong, Haiwen and Huang, Longtao and Xue, Hui and Lu, Weiming and Xiao, Jun and Zhuang, Yueting and Shen, Yongliang},
  year = {2026},
  eprint = {2607.01191},
  archivePrefix = {arXiv},
  primaryClass = {cs.CV},
  url = {https://arxiv.org/abs/2607.01191}
}

\clearpage
\appendix
\setcounter{secnumdepth}{2}
\section{Modular Video-Agent Representation}
\label{app:design-space}

This appendix expands the five-module abstraction introduced in the main paper.

\subsection{Functional Modules and Prior Instantiations}


We revisit the design lineage of long-form video agents and observe that most prior systems, despite substantial implementation differences, can be represented by five recurring functional modules, even when a system fuses multiple functions within one model. \emph{Video structuring} converts raw video into a reusable search substrate, with historical examples spanning frame or clip embedding indexes~\citep{wang2024videoagent}, query-adaptive trees~\citep{wang2025videotree}, dynamic entity-relation graphs~\citep{chu2025graphvideoagent}, segment-level textual databases~\citep{zhang2025deepvideodiscovery}, and entity-centric hierarchical indexes~\citep{yin2026haven}. \emph{Evidence localization} determines how a question retrieves relevant temporal regions from that representation. Historical systems instantiate this role through keyword retrieval~\citep{zhang2024omagent}, embedding-based semantic search~\citep{zhang2025deepvideodiscovery}, learned temporal grounding~\citep{liu2026videomind}, or iterative coarse-to-fine exploration~\citep{yin2025videoarm,lin2026videoseek}. \emph{Perception} turns selected intervals into semantic observations by invoking VLMs and modality-specific extractors such as ASR or OCR~\citep{wang2024videoagent,zhang2024omagent,li2026lenswalk}. \emph{Working memory} accumulates and consolidates observations, tool outputs, temporal indexes, and intermediate reasoning, with designs ranging from hierarchical multimodal memory to separate visual, episodic, and semantic memories~\citep{yin2025videoarm,yeo2025worldmm}. Finally, \emph{reasoning} assesses evidence sufficiency, plans and revises tool calls, and synthesizes the answer, either through a think--act--observe loop or specialized planner--grounder--verifier--answerer roles~\citep{lin2026videoseek,liu2026videomind}. Together, these recurring roles provide a common, interface-stable representation of prior video-agent designs; the examples above merely substantiate this retrospective view and do not exhaust possible implementations or define a closed design space. The modular decomposition itself is not our contribution. MetaVideoAgent instead contributes the automated failure diagnosis and evolution process operating over these modules.

Table~\ref{tab:prior-video-agent-module-mapping} maps the four fixed
video-agent baselines in the main comparison to this abstraction, based on
their released implementations.  The entries denote functional roles rather
than source-code packages: a persistent, question-independent video store is
counted as structuring, whereas retrieved evidence and the state accumulated
while solving a question are counted as working memory.

\begin{table*}[t]
\centering
\scriptsize
\setlength{\tabcolsep}{2pt}
\renewcommand{\arraystretch}{1.15}
\begin{tabular}{p{0.12\textwidth}p{0.164\textwidth}p{0.164\textwidth}p{0.164\textwidth}p{0.164\textwidth}p{0.164\textwidth}}
\toprule
\multicolumn{1}{c}{Method} &
\multicolumn{1}{c}{Video Structuring} &
\multicolumn{1}{c}{Evidence Localization} &
\multicolumn{1}{c}{Perception} &
\multicolumn{1}{c}{Working Memory} &
\multicolumn{1}{c}{Reasoning} \\
\midrule
DVD~\citep{zhang2025dvd}
& Timestamped clip-caption vector database and subject registry
& Semantic clip search and LLM-selected inspection ranges
& VLM inspection over sampled frames in selected ranges
& Message history with tool observations and reflections
& ReAct-style planning, tool selection, and answer finalization \\
M3-Agent~\citep{long2026m3agent}
& 30-second segments and an entity-centric multimodal memory graph
& Iterative embedding retrieval over graph nodes and clips
& Video/audio processing with face and speaker analysis
& Retrieved graph evidence and search dialogue
& Control model alternates between \textsc{Search} and \textsc{Answer} \\
WorldMM~\citep{yeo2025worldmm}
& Multiscale captions with episodic, semantic, and visual stores
& Iterative memory-type and multiscale retrieval
& Frame--transcript captioning and visual feature extraction
& Deduplicated retrieved items and retrieval-round history
& Retrieval reasoning followed by evidence-conditioned QA \\
Symphony~\citep{yan2026symphony}
& Time-ordered frame stream and subtitle source
& Query expansion and relevance-based temporal grounding
& Frame inspection, interval summaries, association, and subtitle analysis
& Shared operation history and reflection feedback
& Core planning, specialist delegation, and reflective answer revision \\
\bottomrule
\end{tabular}
\caption{Functional mapping of the four fixed video-agent baselines in the
main comparison.  The mapping follows the released implementations and
describes functional roles, not a claim that each role is a separately
packaged component.}
\label{tab:prior-video-agent-module-mapping}
\end{table*}

\subsection{Agent Interface and Execution Trace}

Let \(\mathcal{M}=\mathcal{S}\times\mathcal{L}\times\mathcal{P}
\times\mathcal{W}\times\mathcal{R}\) be the modular design space. We
represent an agent as
\begin{equation}
A=(S,L,P,W,R)\in\mathcal{M},
\end{equation}
where \(S\), \(L\), \(P\), \(W\), and \(R\) are video structuring,
evidence localization, perception, working memory, and reasoning modules.
The implementation instantiates every component through a module registry, so
a primary module can be updated while preserving stable interfaces to the
remaining components. When a diagnosed mechanism crosses a module boundary,
the candidate may also make compatible coordinated changes to coupled
components.

Given a video \(V\), the structuring module first constructs a reusable
time-indexed representation \(Z=S(V)\).  Both localization and perception
receive access to \(Z\).  At reasoning step \(t\), the reasoning module reads
the question, video metadata, and memory state \(m_t\), and either terminates
with an answer or emits a tool plan:
\begin{equation}
(h_t,\delta_t,\xi_t)=R(q,V,m_t), \quad
o_t=T_{\delta_t}(\xi_t;Z), \quad T_{\delta_t}\in\{L,P\}.
\end{equation}
Here \(h_t\) is the reasoning record, \(\delta_t\) is a selected tool, and
\(\xi_t\) contains its arguments; when \(R\) emits \textsc{finish}, it
returns an answer instead of invoking a tool. The memory module then updates
\(m_{t+1}=W(m_t,h_t,\delta_t,\xi_t,o_t)\).
This Think--Act--Observe process continues until the agent finishes or reaches
its step budget.  In code, \(R\) is a unified \emph{Thinking} module that
jointly performs planning and answer reasoning.  The execution record stores
thoughts, tool parameters, observations, intermediate variables, the final
answer, and resource statistics.  Consequently, a failure can be attributed to
a concrete module interaction instead of only to the final prediction.


\section{MetaVideoAgent Evolution Algorithms}
\label{app:evolution}

This appendix provides executable pseudocode for the complete MetaVideoAgent
control flow and for each module introduced in Section~\ref{sec:metavideoagent}.
Algorithm~\ref{alg:metavideoagent-full} composes Distribution-Aware Design
(Algorithm~\ref{alg:distribution-aware-design}), Teacher review
(Algorithm~\ref{alg:teacher-review}), minimal-task construction
(Algorithm~\ref{alg:minimal-validation-task}), cross-trajectory diagnosis
(Algorithm~\ref{alg:cross-trajectory-diagnosis}), and responsibility-constrained
code evolution (Algorithm~\ref{alg:responsibility-constrained-evolution}).
Held-out measurements for archived checkpoints are isolated from the evolution
process.  They are retained solely for post-hoc reporting and are never exposed
to review, diagnosis, candidate generation, promotion, stopping, or rollback.

\begin{algorithm}[t]
\caption{Complete MetaVideoAgent Evolution}
\label{alg:metavideoagent-full}
\footnotesize
\begin{algorithmic}[1]
\REQUIRE Evolution split \(\mathcal{D}_{\rm tr}\); held-out split
\(\mathcal{D}_{\rm te}\); optional \(A_{\rm in}\); \(K\) iterations
\ENSURE Agent sequence \(A_0,\ldots,A_K\), history \(\mathcal{H}\), and
held-out record \(\mathcal{E}_{\rm te}\)
\STATE \(A_0,p,h_0,r_0\leftarrow
\mathrm{DistributionAwareDesign}(\mathcal{D}_{\rm tr},A_{\rm in})\)
\STATE \(\mathcal{H}\leftarrow\{p,h_0,r_0\}\)
\STATE \(\mathcal{E}_{\rm te}[0]\leftarrow
\mathrm{ExternalEval}(A_0,\mathcal{D}_{\rm te})\) \COMMENT{record only}
\FOR{\(k=0\) to \(K-1\)}
  \STATE \((\tau_k,s_k)\leftarrow\mathrm{FullEval}(A_k,\mathcal{D}_{\rm tr})\)
  \STATE \(F_k\leftarrow\mathrm{Incorrect}(\tau_k)\)
  \STATE \(\mathcal{R}_k\leftarrow
  \mathrm{TeacherReview}(F_k,Y_{\rm gt},I_{\rm ev},A_k)\)
  \STATE \(Q_k\leftarrow\mathrm{BuildMinimalTasks}(\mathcal{R}_k)\)
  \STATE \(d_k\leftarrow\mathrm{CrossTrajectoryDiagnose}
  (\mathcal{R}_k,Q_k,A_k,\mathcal{H})\)
  \STATE \(A_{k+1},\mathcal{B}_k\leftarrow
  \mathrm{ConstrainedEvolve}(A_k,d_k,Q_k,s_k,\mathcal{D}_{\rm tr})\)
  \STATE \(\mathcal{H}\leftarrow
  \mathrm{UpdateHistory}(\mathcal{H},\tau_k,\mathcal{R}_k,d_k,\mathcal{B}_k)\)
  \STATE \(\mathcal{E}_{\rm te}[k+1]\leftarrow
  \mathrm{ExternalEval}(A_{k+1},\mathcal{D}_{\rm te})\) \COMMENT{no feedback}
\ENDFOR
\RETURN \(A_0,\ldots,A_K,\mathcal{H},\mathcal{E}_{\rm te}\)
\end{algorithmic}
\end{algorithm}

\subsection{Distribution-Aware Design}

\begin{algorithm}[t]
\caption{Distribution-Aware Initial Design}
\label{alg:distribution-aware-design}
\footnotesize
\begin{algorithmic}[1]
\REQUIRE Evolution videos \(\mathcal{V}_{\rm tr}\), associated queries
\(\mathcal{Q}_{\rm tr}\), optional agent \(A_{\rm in}\)
\ENSURE Executable initial agent \(A_0\), profile \(p\), requirements
\(h_0\), and retrieved evidence \(r_0\)
\STATE \(p\leftarrow\emptyset\)
\FOR{each video \(V_i\in\mathcal{V}_{\rm tr}\)}
  \STATE \(F_i\leftarrow\mathrm{UniformSample}(V_i,5)\)
  \STATE \(s_i\leftarrow\mathrm{Summarize}
  (F_i,\mathrm{Metadata}(V_i),\mathcal{Q}_i)\)
  \STATE \(p\leftarrow p\cup\{s_i\}\) \COMMENT{no answers or evidence labels}
\ENDFOR
\STATE \(h_0\leftarrow\mathrm{InduceRequirements}(p)\)
\STATE \(r_0\leftarrow\mathrm{RetrieveSources}(h_0)\)
\STATE \(A_{\rm seed}\leftarrow A_{\rm in}\) if valid; otherwise
\(A_{\rm seed}\leftarrow\emptyset\)
\STATE \(A_0\leftarrow\mathrm{InitDesign}(A_{\rm seed},p,h_0,r_0)\)
\STATE \(A_0\leftarrow\mathrm{RepairAndSmoke}(A_0)\)
\RETURN \(A_0,p,h_0,r_0\)
\end{algorithmic}
\end{algorithm}

\subsection{Teacher Review and Minimal Validation}

\begin{algorithm}[t]
\caption{Teacher Video Agent Question-Level Review}
\label{alg:teacher-review}
\footnotesize
\begin{algorithmic}[1]
\REQUIRE Incorrect Student trajectory \(\tau_{k,i}\), question \(q_i\),
answer \(y_i\), evidence interval \(I_i^{\rm ev}\), agent \(A_k\)
\ENSURE Review \(\rho_{k,i}\)
\STATE \(e_i\leftarrow\mathrm{EvidenceView}(V_i,I_i^{\rm ev})\)
\STATE \(z_i\leftarrow\mathrm{InspectStructuring}(A_k,I_i^{\rm ev})\)
\STATE \(\gamma_i\leftarrow\mathrm{BuildGoldPath}(q_i,y_i,e_i,z_i)\)
\STATE \(t_i^\star\leftarrow\textsc{FinalDecision}\) \COMMENT{fallback attribution}
\FOR{each Student step \(t\) in temporal execution order}
  \STATE \(c_t\leftarrow\mathrm{CanSupportAnswer}
  (\tau_{k,i}^{\leq t},\gamma_i)\)
  \IF{\(c_t=\textsc{false}\)}
    \STATE \(t_i^\star\leftarrow t\); \textbf{break}
  \ENDIF
\ENDFOR
\STATE \(\phi_{k,i}\leftarrow\mathrm{InferCause}
(\tau_{k,i},\gamma_i,t_i^\star,e_i,z_i)\)
\STATE \(m_{k,i}\leftarrow\mathrm{AttributeModule}
(\phi_{k,i},A_k)\)
\STATE \(v_{k,i}\leftarrow\mathrm{ProposeMinimalTask}
(\phi_{k,i},m_{k,i},e_i)\)
\RETURN \(\rho_{k,i}=(t_i^\star,\phi_{k,i},m_{k,i},v_{k,i})\)
\end{algorithmic}
\end{algorithm}

\begin{algorithm}[t]
\caption{Minimal Validation Task Construction}
\label{alg:minimal-validation-task}
\footnotesize
\begin{algorithmic}[1]
\REQUIRE Reviews \(\mathcal{R}_k=\{\rho_{k,i}\}_{i\in F_k}\)
\ENSURE Consolidated validation set \(Q_k\)
\STATE \(U_k\leftarrow\emptyset\)
\FOR{each review \(\rho_{k,i}\in\mathcal{R}_k\)}
  \STATE \((z_{k,i},y_{k,i}^{\rm exp})\leftarrow
  \mathrm{IsolateFailure}(\rho_{k,i})\)
  \IF{\(\mathrm{IndependentExecutable}(z_{k,i})\land
  \mathrm{ReliableTarget}(y_{k,i}^{\rm exp})\)}
    \STATE \(u_{k,i}\leftarrow(z_{k,i},y_{k,i}^{\rm exp})\)
  \ELSE
    \STATE \(u_{k,i}\leftarrow\mathrm{OriginalVideoQA}(i)\)
  \ENDIF
  \STATE \(U_k\leftarrow U_k\cup\{u_{k,i}\}\)
\ENDFOR
\STATE \(Q_k\leftarrow\mathrm{ClusterDeduplicateAndCover}
(U_k,\mathcal{R}_k)\)
\RETURN \(Q_k\)
\end{algorithmic}
\end{algorithm}

\subsection{Cross-Trajectory Diagnosis}

\begin{algorithm}[t]
\caption{Diagnosis Agent Cross-Trajectory Diagnosis}
\label{alg:cross-trajectory-diagnosis}
\footnotesize
\begin{algorithmic}[1]
\REQUIRE Reviews \(\mathcal{R}_k\), validation tasks \(Q_k\), current
agent \(A_k\), history \(\mathcal{H}_k\)
\ENSURE Diagnosis contract \(d_k=(m_k,\phi_k,Q_k,\Gamma_k,\Pi_k)\)
\STATE \(C\leftarrow\mathrm{ClusterByCauseAndModule}(\mathcal{R}_k)\)
\STATE \(C\leftarrow\mathrm{SeparateEngineeringAndDesignFailures}(C)\)
\STATE \(C\leftarrow\mathrm{ScoreRecurrenceImpactAndEvidence}(C,Q_k)\)
\STATE \(C\leftarrow\mathrm{PenalizeRepeatedDirections}(C,\mathcal{H}_k)\)
\IF{evidence is insufficient or conflicting}
  \STATE \(a_k\leftarrow\mathrm{TargetedEvidence}(C,A_k,\mathcal{H}_k)\)
  \STATE \(C\leftarrow\mathrm{UpdateEvidence}(C,a_k)\)
\ENDIF
\STATE \((m_k,\phi_k)\leftarrow\mathrm{SelectPrimaryFailure}(C)\)
\STATE \(\Gamma_k\leftarrow\mathrm{SetScopeAndInterfaceConstraints}(m_k,A_k)\)
\STATE \(\Pi_k\leftarrow\mathrm{SetValidationAndCostPlan}
(\phi_k,Q_k)\)
\RETURN \(d_k=(m_k,\phi_k,Q_k,\Gamma_k,\Pi_k)\)
\end{algorithmic}
\end{algorithm}

\subsection{Responsibility-Constrained Code Evolution}

\begin{algorithm}[t]
\caption{Evolution Agent Update and Candidate Acceptance}
\label{alg:responsibility-constrained-evolution}
\footnotesize
\begin{algorithmic}[1]
\REQUIRE Current agent \(A_k\), diagnosis \(d_k\), tasks \(Q_k\),
baseline evidence \(s_k\), evolution split \(\mathcal{D}_{\rm tr}\)
\ENSURE Accepted agent \(A_{k+1}\) and candidate artifacts \(\mathcal{B}_k\)
\STATE \(A_{k+1}\leftarrow A_k\)
\IF{\(\mathrm{Decision}(d_k)\neq\textsc{Evolve}\)}
  \RETURN \(A_{k+1},\{d_k\}\)
\ENDIF
\STATE \(\widetilde A_k\leftarrow\mathrm{ModifyPrimaryModule}(A_k,d_k)\)
\STATE \(\widetilde A_k\leftarrow\mathrm{ModifyNecessaryDependencies}
(\widetilde A_k,\Gamma_k)\)
\STATE \((\widetilde A_k,T_k)\leftarrow
\mathrm{RepairAndSmoke}(\widetilde A_k)\)
\STATE \(G_k^{\rm eng}\leftarrow\mathrm{EngineeringGate}(T_k)\)
\IF{\(G_k^{\rm eng}=0\)}
  \RETURN \(A_{k+1},\{\widetilde A_k,T_k\}\)
\ENDIF
\STATE \(P_k\leftarrow\mathrm{ProbeAndAgentReplay}(\widetilde A_k,Q_k)\)
\STATE \(G_k^{\rm probe}\leftarrow\mathrm{ProbeGate}(P_k)\)
\IF{\(G_k^{\rm probe}=0\)}
  \RETURN \(A_{k+1},\{\widetilde A_k,T_k,P_k\}\)
\ENDIF
\STATE \(\widetilde s_k\leftarrow
\mathrm{FullEval}(\widetilde A_k,\mathcal{D}_{\rm tr})\)
\IF{\(\mathrm{Accept}(\widetilde s_k,s_k)\)}
  \STATE \(A_{k+1}\leftarrow\widetilde A_k\)
\ENDIF
\STATE \(\mathcal{B}_k\leftarrow
\{\widetilde A_k,T_k,P_k,\widetilde s_k,A_{k+1}\}\)
\RETURN \(A_{k+1},\mathcal{B}_k\)
\end{algorithmic}
\end{algorithm}

\subsection{Review, Diagnosis, and Validation Details}
\label{app:review-validation-details}

The Teacher Video Agent uses two constrained review tools.
\textsc{EvidenceView} revisits an annotated temporal interval with an
adjustable sampling rate and perception model, whereas
\textsc{InspectStructuring} retrieves the Student's stored content aligned
with that interval. Question-level reviews are the primary evidence for
Diagnosis. When necessary, the Diagnosis Agent additionally inspects
representative trajectories and history, revisits disputed evidence, retrieves
source-grounded methods for the responsible module, or validates a proposed
model, API, tool, or local algorithm on a minimal input. Historical comparison
does not mechanically blacklist a previously attempted direction: it compares
the observed effects and interactions of archived evolution-split designs and
trajectories. Codegen therefore receives the current accepted agent, the
Diagnosis contract, and selected historical references, while every generated
code version and validation trace is retained in the evolution history.
Static checking and smoke execution repair or reject invalid code before
minimal validation; the engineering decision is based on interface and runtime
trace validity, not on whether the smoke question is answered correctly. A candidate
passing these local checks is still promoted only after full evolution-split
comparison.  The probe is an iterative evolution driver rather than a
one-shot rejection gate: when trace analysis finds that the intended
capability has not improved on the minimal-validation set, the execution trace
and observed answer behavior are returned to Codegen to guide another code
revision.  The revised bundle repeats repair, smoke execution, and the same
probe until trace-based analysis verifies the intended improvement.  Thus, the
probe stage does not define a separate failure exit; an unimproved result
remains actionable feedback for Codegen. Exiting this loop denotes
\textsc{Proceed} and satisfies the minimal-validation gate.

\paragraph{A concrete Gold-Path construction.}
Consider an Interview question asking how many fruits hung from the branches
in a small vase when the speaker said, ``In fact, the earlier shape was this
kind of whisk.''  The Teacher first decomposed the question into a temporal
anchor and a visual counting obligation.  Within the annotated evidence
interval, ASR and subtitles aligned the quoted sentence with the end of the
segment, at approximately 737\,s.  They established when to look, but did not
provide the requested count.  The Teacher then inspected multiple wide-view
frames spanning the interval.  The same vase remained visible on the left side
of the table, with three orange fruits---one above and two below---hanging from
its branches.  The resulting Gold Path was therefore: ground the quoted
utterance, identify the referenced vase in the concurrent scene, verify a
stable count across frames, and answer \emph{three}.  Visual evidence was the
primary answer channel, while speech and subtitles served only as temporal
anchors.

This example illustrates that a Gold Path is neither a prescribed Student
tool sequence nor a rationale inferred from the answer label alone.  It is a
minimal sufficient evidence chain independently reconstructed by the Teacher
from the question, source answer, annotated evidence interval, and raw media.
The Student's aligned video structure is inspected only afterward, when the
independently constructed chain is used to audit the trajectory and attribute
its first causal divergence.  The chain specifies which facts must hold,
which channels establish them, and how they entail the answer.  In this case,
the Student produced mutually
inconsistent counts and finally selected four, although no observation in its
trajectory supported that number.  Source answers and evidence intervals are
available only to Teacher review on the evolution split; the executable
Student does not receive either annotation.

\subsection{A Worked Evolution Cycle}
\label{app:worked-evolution-cycle}

\begin{figure*}[p]
\centering
\includegraphics[width=\textwidth,height=0.82\textheight,keepaspectratio]{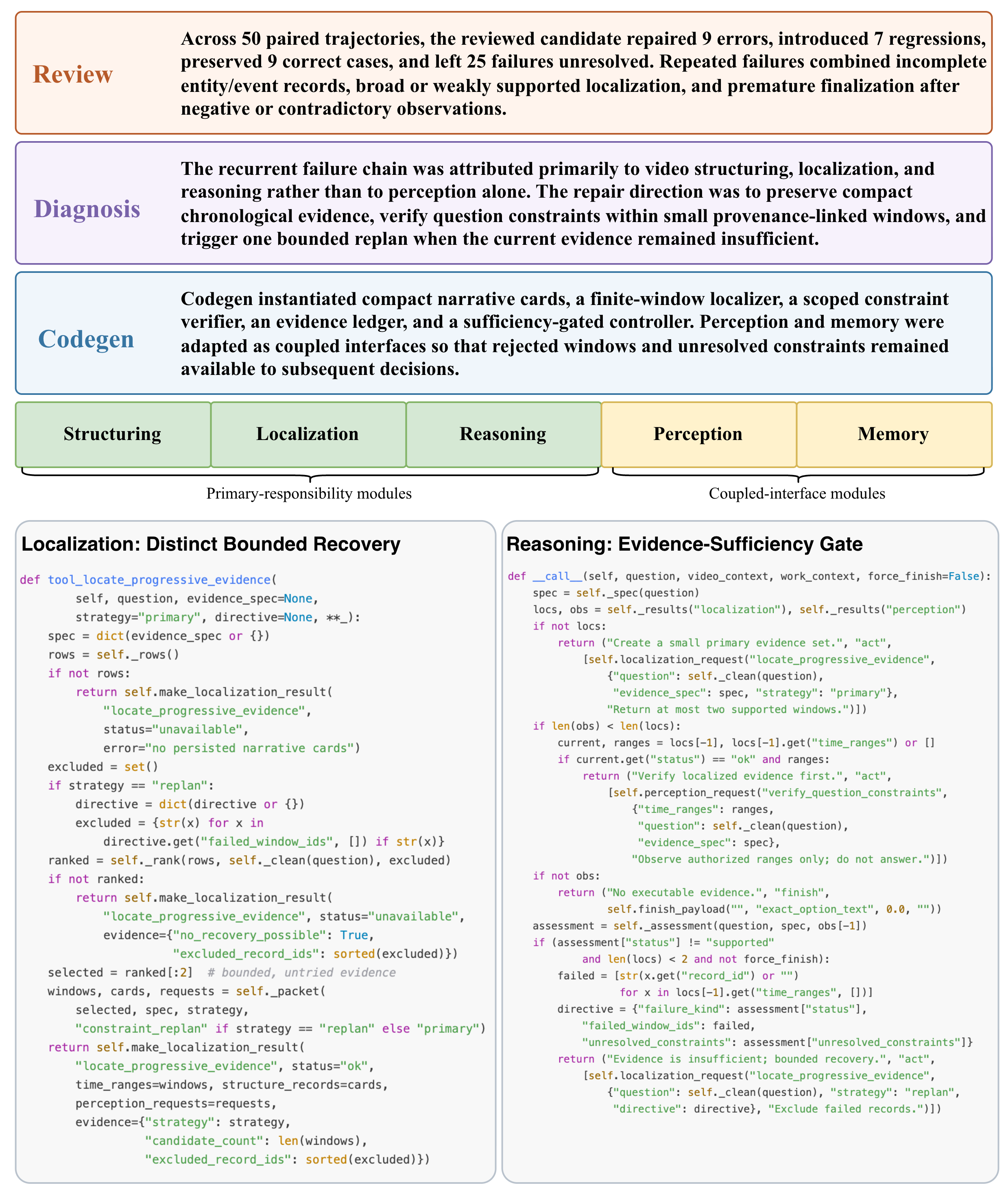}
\caption{A worked Review--Diagnosis--Codegen cycle on Dramatic Narrative. Review identifies recurrent failures in the evidence flow; Diagnosis assigns primary responsibility to video structuring, localization, and reasoning; and Codegen also adapts the coupled perception and memory interfaces required to execute the repair. The code panels show selected excerpts from the generated candidate that preserve the control logic relevant to bounded recovery and evidence-sufficiency gating. All diagnostic evidence is drawn from the evolution split.}
\label{fig:review-diagnosis-codegen}
\end{figure*}

Figure~\ref{fig:review-diagnosis-codegen} summarizes one actual Dramatic Narrative update.  Its prose is an author-level reconstruction grounded in the
archived evolution-split trajectories, Teacher review, Diagnosis contract, and
generated bundle, rather than a verbatim reproduction of prompts or JSON.
This reconstruction makes explicit the design meaning that is distributed
across otherwise lengthy execution artifacts.

\paragraph{Review.}
The review compared 50 paired trajectories, recording nine corrections, seven
regressions, nine preserved correct cases, and 25 persistent failures.  These
outcomes showed that the candidate contained useful behavior but was not a
uniform improvement: some additional inspection happened to repair individual
questions, while related questions still failed or regressed.  Review
therefore examined how evidence moved through the agent instead of treating
the net score as a sufficient explanation.

Two surface-different failures expose the recurrent pattern.  One question
requires finding when a brief action involving a blue-braided character occurs.
The earlier agent searched broadly related plot material but did not isolate
the transient event as a supported temporal observation.  Another question
asks for the spatial relation between two characters when a quoted subtitle is
spoken.  Relevant material entered the evidence path, but the observed
relation was not reliably converted into the semantic relation required by the
answer.  Across these and other failures, coarse or incomplete entity--event
records led to weakly supported localization; negative or contradictory
perception did not alter the plan; and reasoning could still terminate.

This pattern also explains why the review did not assign the defect to
perception alone.  In several trajectories, a relevant region was already
among the candidates, and broader or repeated perceptual inspection still did
not produce a grounded decision.  The missing behavior lay both upstream and
downstream of perception: the agent lacked an explicit representation of what
had to be verified, and it lacked a control rule that converted insufficient
observations into a directed new search.  Review therefore interpreted the
failures as a recurrent evidence-control defect rather than unrelated VLM
mistakes.

\paragraph{Diagnosis.}
Diagnosis converted this cross-trajectory pattern into an executable design
requirement.  Each question would first be represented by an evidence
specification containing the target entities or actions, relevant temporal or
spatial relation, required evidence channels, and currently unresolved
constraint.  Video structuring had to expose compact chronological records
with corresponding entity, action, and relation fields.  Localization had to
return a small set of finite, provenance-bearing windows.  Reasoning then had
to test whether scoped observations supported the specification before it
could answer.

The recovery was deliberately bounded.  The reviewed candidate often reacted
to uncertainty by inspecting more broadly, which increased cost and introduced
similar but irrelevant scenes without clarifying which hypothesis was being
tested.  Diagnosis instead required a failed window to be recorded and
excluded, and allowed a new localization request only for a named unresolved
constraint.  This turns recovery from a global rescan into a test of a specific
missing piece of evidence.

\paragraph{Codegen.}
Codegen realized this requirement as a connected bundle.  Structuring produced
compact narrative cards; localization ranked those cards and returned at most
two supported windows; perception observed only the authorized ranges; memory
retained accepted, rejected, and unresolved evidence; and reasoning applied an
evidence-sufficiency gate.  The left code excerpt in Figure~\ref{fig:review-diagnosis-codegen} shows that recovery consumes the
failed record identifiers, removes them from the next ranking, and reports
that no recovery is possible when no distinct evidence remains.  Selecting
only the top two remaining records prevents the recovery path from silently
becoming another global scan.

The right excerpt implements the complementary consumer-side rule.  An
answer-shaped string is not treated as evidence: if the latest observations do
not support the evidence specification, reasoning packages the failure type,
failed windows, and unresolved constraints into one bounded re-localization
request.  Hence the code changes form a closed interaction---localization
proposes evidence, perception verifies it, reasoning checks sufficiency, and
memory carries the failed hypothesis into a distinct retry---rather than five
independent prompt edits.

\paragraph{Minimal-validation probe.}
Minimality is applied at two complementary levels.  At the task level, when
the review evidence permits reliable isolation, a failure from the original
long-video QA execution is converted into an independently executable probe
over the smallest sufficient time-localized multimodal context.  Its expected
target is used to assess the probe execution rather than supplied to the
candidate.  When such isolation would not produce a reliable executable task,
the original video-QA instance is retained as a fallback.  At the set level,
Diagnosis clusters and deduplicates the resulting localized tasks and fallback
instances into a small set focused on the diagnosed mechanism.  Thus,
``minimal'' does not require every task to be a clip-level question.

Question-level reviews proposed candidate witnesses, and Diagnosis
deduplicated them into a five-task minimal-validation set.  Two were direct
repair witnesses: one tested whether the agent could isolate and verify a
short-lived plot event, and the other tested whether verified spatial evidence
was consumed as a semantic relation.  Three adjacent guards covered visual
appearance, fine-grained action, and narrative causality.  The guards were not
expected to represent the same defect; they checked whether the new evidence
loop merely overfit the two target forms or disturbed neighboring behavior.

The second direct repair witness makes the role of a minimal-validation task
concrete.  It asks where the male protagonist is relative to a man wearing a
yellow helmet when a quoted subtitle is spoken.  The earlier combo observed
that the two men were riding a scooter and described the protagonist both as
being on the image's right side and as sitting behind the helmeted driver, but
reasoning copied the image-plane direction \emph{right} into the answer.  Review
isolated the failure as a coordinate-frame mismatch: the question requests a
scene-relative relation, whereas the camera's side view supplies an
image-relative position.  Diagnosis therefore required the candidate to keep
the target and reference entities, their scene roles, and their spatial
relation distinct, and to resolve the requested coordinate frame before
mapping evidence to an option.

This archived witness used the permitted full-video-QA fallback rather than
task-level media compression.  The original question and video context were
retained as the executable probe, while the candidate received neither the
answer \emph{rear} nor the annotated evidence interval.  It therefore
illustrates diagnosis-specific mechanism verification under the fallback; its
cost reduction comes from set-level selection, whereas isolated tasks can
additionally reduce per-probe cost through scoped media inputs.  Its executed
trajectory identified the yellow-helmeted man as the driver and the protagonist
as the passenger, retained the side-view observation without treating it as
the requested relation, and concluded that the protagonist was behind the
driver.  Thus, the validated behavior was not a narrow rule that ``right means
rear'' on this video.  It was the reusable ability to distinguish image-plane
coordinates from scene-relative relations using entity roles and scene
structure.  Had the candidate again copied a screen direction without this
conversion, the trajectory would have supplied a concrete repair signal and
returned the candidate to Codegen.  This direct repair witness is therefore an
executable, falsifiable capability test within the minimal-validation set, not
an assumption that generated code will recover a Teacher-provided answer.

On both direct witnesses, the candidate executed the intended
localize--verify--assess path and corrected the earlier error.  The guard
trajectories provided an equally important boundary.  The agent could reach
the relevant scene yet still misread weather appearance, confuse a character's
action, or infer the wrong narrative cause.  These failures showed that the
new bundle repaired evidence localization and sufficiency control, but did not
by itself solve every perceptual or causal-reasoning problem.  The unresolved
guard traces therefore remained revision evidence rather than being hidden by
the two successful repairs.

The minimal set is not a performance estimate or a held-out validation split.
Its purpose is to test a diagnosis-specific mechanism, including the concrete
module calls and state transitions, before repeatedly paying for a full
evolution-split comparison.  Whenever reliable, task-level compression reduces
the media context needed by an individual probe; fallback tasks retain the
original video-QA obligation when such isolation is unreliable.  Across both
forms, set-level compression avoids rerunning every heterogeneous
evolution-split error after each code edit, which would mix unrelated failure
causes and obscure whether the intended handoff had changed.  The resulting
small set shortens the design--execution--inspection loop while preserving the
full agent trajectory needed for the next repair.

\subsection{Selective Modular Evolution}
\label{app:selective-modular-evolution}

\begin{figure*}[p]
\centering
\includegraphics[width=\textwidth,height=0.82\textheight,keepaspectratio]{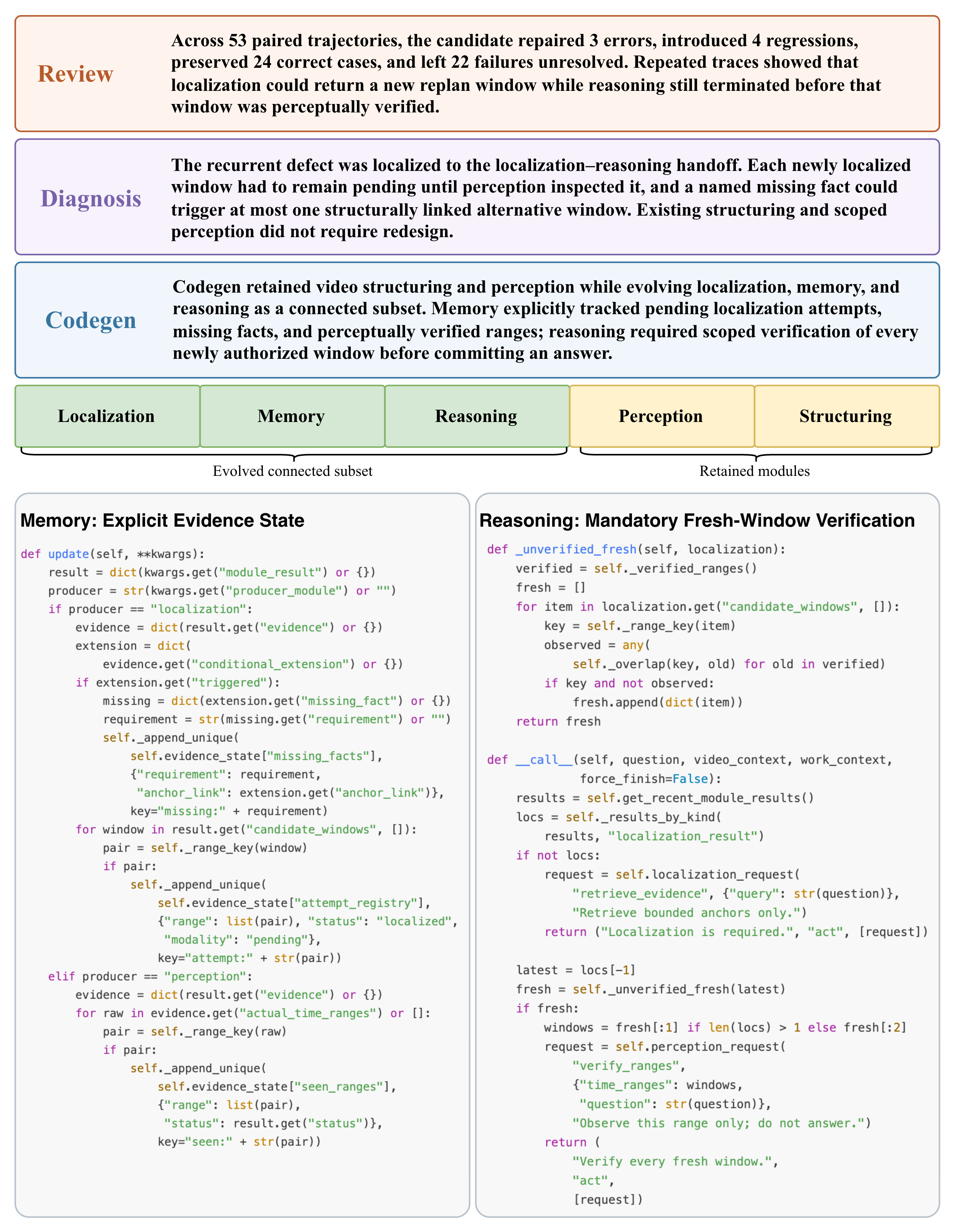}
\caption{A selective modular-evolution example on Product Presentation. Review identifies premature termination before a newly localized window is perceptually verified. Diagnosis isolates the localization--reasoning handoff, and Codegen evolves the connected localization--memory--reasoning subset while retaining the functional structuring and perception modules. The code panels show selected excerpts relevant to explicit evidence-state tracking and mandatory fresh-window verification.}
\label{fig:selective-modular-evolution}
\end{figure*}

Figure~\ref{fig:selective-modular-evolution} shows a complementary Product Presentation update in which the diagnosed failure did not require replacing
the complete bundle.

\paragraph{Review.}
Across 53 paired trajectories, the candidate repaired three errors, introduced
four regressions, preserved 24 correct cases, and left 22 failures unresolved.
The informative pattern appeared after re-localization: a new candidate window
could be returned, yet reasoning could finish before perception had observed
that fresh window.  Consequently, generating an alternative location did not
guarantee that the alternative contributed evidence to the answer.  The
already available product index and scoped perception operation were not the
source of this state-transition failure.

\paragraph{Diagnosis.}
Diagnosis therefore isolated the localization--reasoning handoff.  A fresh
window had to remain explicitly pending until perception observed it, and
reasoning had to check for such pending evidence before completion.  To keep
the recovery directed, at most one structurally linked alternative could be
requested for one named missing fact, such as an unseen object view or
attribute.  This diagnosis preserved video structuring and perception while
requiring a state-bearing interface between localization and reasoning.

\paragraph{Codegen.}
Codegen retained the existing structuring and perception modules and evolved
localization, working memory, and reasoning.  Localization associated an
alternative window with the named missing fact.  The memory excerpt in Figure~\ref{fig:selective-modular-evolution} explicitly tracks pending
localization attempts, missing facts, and perceptually verified ranges.  The
reasoning excerpt queries this state for a fresh unverified range and forces a
scoped perception call before allowing the agent to finish.  The example
therefore illustrates both sides of modular evolution: MVA can jointly update
the modules forming a failed handoff, and it can preserve modules for which the
review provides no reason to redesign.

\paragraph{Stability of modular evolution.}
Here, stability does not mean that only one module may change or that accuracy
must increase monotonically at every archived checkpoint.  Review and
Diagnosis bound an update by a causal failure chain, after which Codegen may
modify the connected subset needed to repair that chain while retaining the
current best's functioning behavior and interfaces elsewhere.  Consequently,
cross-module changes are permitted when a producer--consumer handoff requires
them, but an unrelated bundle-wide rewrite is not the default response to a
local failure.

Stability is further enforced through explicit interface states and staged
execution evidence.  A newly localized range, for example, progresses from
\texttt{localized} to \texttt{pending} and only then to \texttt{verified} or
\texttt{failed}; reasoning cannot treat pending evidence as observed, and
memory preserves verified, rejected, and unresolved facts across steps.
Static checks and smoke execution first expose source, interface, and runtime
faults.  Diagnosis-specific probes then test whether the intended capability
and state transition actually occur in the complete agent, together with
adjacent guards that reveal immediate regressions.  Passing a probe does not
by itself promote the candidate: the subsequent full evolution-split
comparison records corrections, regressions, and unresolved failures before
the current best can be replaced.  Modular evolution is thus stabilized by
diagnosis-bounded scope, interface invariants, trajectory-level probes, and
full-split regression evidence rather than by assuming that a local edit is
inherently safe.

Together, the two cases show that MVA evolves distribution-relevant control
mechanisms rather than adding rules for individual questions.  The Dramatic
Narrative case requires a bundle-wide evidence loop, whereas the Product
Presentation case repairs a narrower state transition by changing only the
connected subset responsible for it.

\section{Experimental Setup and Implementation Details}
\label{app:experimental-setup}
\label{app:implementation}

\paragraph{Independent MVA runs.}
We run MVA independently from scratch for every target distribution.  A run
begins with Distribution-Aware Design: sparse observations of the evolution
videos and the associated query set yield a distribution profile used for
initial research and code generation.  The initial research stage is mandatory;
later targeted research is invoked only when the diagnosis requires additional
evidence.  All runs use the same five-module design space and the same
review--diagnosis--code-evolution--validation protocol.  Profiles, code,
trajectories, review artifacts, and evolved designs are never shared across
target distributions. The main results report these from-scratch runs; the
DVD-initialized setting is used only for the Course Lecture ablation.

\paragraph{Model and tool configuration.}
The Teacher and Diagnosis Agents use Qwen 3.7 Max; the Student reasoning module
uses GLM-5.2; and the code-level Evolution Agent uses Codex.  Teacher and
Student invoke Qwen 3.7 Plus for VLM perception, Qwen-VL-OCR for OCR,
Qwen3-ASR-Flash for ASR, and Qwen3-VL-Embedding for embedding-based retrieval.
GLM-5.2, Qwen 3.7 Max, and Qwen 3.7 Plus are all run without thinking mode.

\paragraph{Unified video-agent API stack.}
To control backend capability in the main comparison, MVA, DVD, M3-Agent,
WorldMM, and Symphony use the same API assignments whenever their respective
workflows invoke the corresponding capability. Their control flows, module
organization, and tool-selection policies remain unchanged.
\begin{table}[t]
\centering
\scriptsize
\begin{tabular}{ll}
\toprule
Capability & API assignment \\
\midrule
LLM tool call & Qwen 3.7 Max \\
VLM perception & Qwen 3.7 Plus \\
OCR & Qwen-VL-OCR \\
ASR & Qwen3-ASR-Flash \\
Embedding retrieval & Qwen3-VL-Embedding \\
\bottomrule
\end{tabular}
\caption{Unified API assignments for MVA and the four fixed video agents in
Table~\ref{tab:main_results}.}
\label{tab:unified-api-stack}
\end{table}
For MVA itself, the Student reasoning module remains GLM-5.2, while the Teacher
and Diagnosis Agents use Qwen 3.7 Max and the code-level Evolution Agent uses
Codex. Thus, the table standardizes callable video-understanding tools without
replacing each agent's reasoning and control architecture.

\subsection{Initial and Final Distribution-Adaptive Designs}

Each initial agent is synthesized from its own distribution profile and query
set rather than instantiated from a shared hand-written agent.  All eight
initial designs follow the same functional contract: video structuring stores
timestamped multimodal evidence; localization proposes question-conditioned
media ranges; perception inspects only those ranges; working memory preserves
the evidence and its provenance across steps; and the reasoning module decides
whether to gather more evidence or produce the final answer.  The profile
determines how these functions are instantiated, including which evidence
channels are emphasized and how retrieval and verification are coordinated.
It does not assign hand-written rules to individual questions.

\begin{table*}[t]
\centering
\small
\setlength{\tabcolsep}{5pt}
\begin{tabular}{p{0.15\textwidth}p{0.79\textwidth}}
\toprule
Target distribution & Profile-conditioned functional design of the initial agent \\
\midrule
Interview &
Builds an audio-first, timestamped multimodal evidence record; retrieves a
small set of question-relevant intervals; verifies them with scoped ASR and
visual inspection; and carries the verified evidence through a bounded
retrieve--verify--answer loop. \\
Product Presentation &
Organizes speech, visible product attributes, actions, and on-screen text as
temporal product evidence; retrieves candidate product moments before scoped
audiovisual inspection; and aggregates the resulting observations for the
reasoning module. \\
Software Tutorial &
Aligns narration, interface text, and visible operations in timestamped
records; localizes query-relevant interface windows; verifies them through
scoped speech and visual perception; and supports iterative evidence gathering
when the current observation is insufficient. \\
Dramatic Narrative &
Maintains chronological dialogue, action, object, and scene evidence; applies
coarse-to-fine temporal retrieval; inspects the returned intervals with scoped
multimodal tools; and preserves provenance and conflicting observations for
evidence-first reasoning. \\
Stage Performance &
Records time-bounded stage actions and available speech or visual cues;
retrieves a compact set of candidate performance intervals; performs scoped
visual verification; and retains the resulting evidence for a bounded final
decision. \\
Gameplay &
Uses action-centered temporal records augmented with speech and screen-text
signals; retrieves a few candidate gameplay intervals; verifies them with
scoped multimodal perception; and executes a bounded
retrieve--verify--synthesize loop. \\
Sports Broadcast &
Builds timestamped transcript-and-visual evidence for long sports videos;
localizes a small set of query-relevant moments; applies scoped media
inspection; and preserves the complete evidence trajectory for the final
reasoning decision. \\
Course Lecture &
Uses speech and screen text as the primary temporal evidence, retrieves
bounded lecture intervals, visually verifies the localized material, and
retains provenance while synthesizing the answer from the collected evidence. \\
\bottomrule
\end{tabular}
\caption{Functional designs synthesized for the eight initial MVA agents.
Descriptions are given at the capability level rather than as
implementation-specific module or class names.}
\label{tab:initial-functional-designs}
\end{table*}

These initial agents provide distribution-conditioned evidence-access
strategies, but they do not already contain the specialized behaviors used in
our qualitative examples.  Event ordering for ordinal stage questions,
object-relative position reasoning in product videos, and cross-view subject
tracking in sports videos are induced by subsequent evolution.

\paragraph{Final evolved designs.}
Table~\ref{tab:final-functional-designs} summarizes the adaptive functionality
of the final agent produced for each target distribution.  As in
Table~\ref{tab:initial-functional-designs}, the descriptions are stated at the
capability level: they characterize the resulting evidence-access and control
behavior rather than exposing experiment-specific class names.

\begin{table*}[t]
\centering
\scriptsize
\setlength{\tabcolsep}{5pt}
\renewcommand{\arraystretch}{1.08}
\begin{tabular}{p{0.15\textwidth}p{0.79\textwidth}}
\toprule
Target distribution & Distribution-adaptive functionality of the final evolved agent \\
\midrule
Interview &
Organizes the video as a speech-centered multimodal timeline and performs
global-to-local retrieval using quoted content, speakers, and visual anchors.
It prioritizes scoped ASR for statements and intent, invokes visual or OCR
verification for identities, actions, and displayed text when needed, and
continues gathering evidence until the remaining answer alternatives can be
distinguished. \\
Product Presentation &
Represents product segments through products, attributes, demonstrations,
orientation, and speech anchors.  It preserves direct product matches while
adding only locally related views when an attribute remains unresolved, then
verifies the requested side, object relation, visible text, or spoken property
before committing to an answer. \\
Software Tutorial &
Aligns narration, interface text, visible operations, and state transitions in
a provenance-bearing tutorial index.  It forms question-conditioned candidate
windows, verifies the relevant interface state with scoped visual, OCR, or ASR
evidence, and performs a bounded replan when the current observation does not
establish the requested operation or result. \\
Dramatic Narrative &
Builds progressive narrative cards that retain scenes, characters, actions,
relations, dialogue, and their temporal provenance.  It tests whether localized
observations jointly satisfy the question's character--event constraints and,
when they do not, performs a bounded recovery with a distinct narrative anchor
before answering from verified evidence. \\
Stage Performance &
Maintains an event-centered record of performers, props, actions, transitions,
and temporal relations.  For ordinal or relation-dependent questions, it can
establish a global event ordering, retrieve the relevant performance phase,
and verify the selected event locally, rather than treating repeated stage
scenes as interchangeable. \\
Gameplay &
Constructs a dense state timeline containing characters, objects, actions,
HUD information, viewpoints, and before--after transitions.  It routes each
question to bounded state or event windows, observes them in chronological
overview--refine--confirm stages, and uses accumulated state evidence to decide
whether to inspect another route or answer. \\
Sports Broadcast &
Combines timestamped broadcast evidence with subject and identity anchors
across camera views.  It preserves which observations belong to the tracked
athlete, separates observed from unresolved attributes, and uses compact
follow-up inspection to verify attire, actions, scores, or other
temporally localized sports evidence. \\
Course Lecture &
Represents lecture speech, screen text, visual state, layout, and slide
transitions as complementary evidence atoms.  It routes semantic questions to
ASR, textual and layout questions to OCR-supported inspection, and operation or
state-change questions to ordered visual comparison, with bounded
re-localization when a required evidence atom is missing. \\
\bottomrule
\end{tabular}
\renewcommand{\arraystretch}{1.0}
\caption{Distribution-adaptive functional designs of the eight final evolved
MVA agents.  Descriptions are given at the capability level rather than as
implementation-specific module or class names.}
\label{tab:final-functional-designs}
\end{table*}

The resulting agents do not converge to one universal fixed pipeline.
Interview emphasizes speech-centered retrieval with selective visual
verification; Product Presentation and Software Tutorial resolve object views,
interface states, and local relations; Dramatic Narrative and Stage
Performance organize temporally related events; Gameplay follows dynamic state
transitions; Sports Broadcast preserves subject identity across views; and
Course Lecture composes speech, text, layout, and visual change as distinct
evidence atoms.  These adaptations correspond to recurring evidence patterns
of their target distributions rather than rules tied to individual questions.
Section~\ref{app:qualitative} traces representative evolved behaviors through
their executed evidence trajectories.

\subsection{Baseline Inputs and Resource Accounting}

A direct MLLM receives uniformly sampled frames from the complete video,
together with the current question and its options, in one model call.  It has
no ASR, OCR, external tools, or separate video database.  The frame counts used
for each direct MLLM are reported in Table~\ref{tab:main_results}.  DVD uses
the cost-reduced configuration (1 FPS indexing, top-$32$ clip retrieval, and
at most 12 frames in one retrieval), with Qwen 3.7 Max as its LLM, Qwen 3.7
Plus as its VLM, and Qwen3-VL-Embedding for retrieval.  MVA, DVD, M3-Agent,
WorldMM, and Symphony use the unified API tool stack described above while
retaining their respective agent control flows.  All baselines are evaluated on
the same target splits as MVA.

Tok./q and Frm./q report the per-question token and image-frame statistics for
every method. The reported values use the common accounting implementation for
the main comparison.

\section{Dataset and Evaluation Protocol}
\label{app:evaluation}

\subsection{Target Video Distributions}
\label{app:target-video-distributions}

We derive VA-EvoBench from CG-Bench video--QA annotations and instantiate the
evolution protocol on eight video distributions.
Each distribution has an independent evolution split and held-out split; an
agent is never evolved jointly across distributions.  Table~\ref{tab:split-statistics}
reports the question counts.  During distribution profiling, the system sees
the evolution videos and their query set, but not answer labels.  Answer labels
and evidence intervals are exposed only to Teacher review after the Student has
completed its evolution-split trajectory.

\paragraph{Construction and released manifests.}
We form each target distribution by manually grouping CG-Bench videos with
recurring content forms, modality compositions, and evidence-localization
requirements.  Curators inspect ten uniformly spaced frames over the full
duration of every candidate video before assigning it to a distribution.  The
public \textsc{VA-EvoBench} release provides the exact question membership used
in our experiments, rather than an unfiltered source-annotation superset.  It
contains separate \texttt{evolution.jsonl} and \texttt{held\_out.jsonl}
manifests, a video-level split manifest, and the construction criteria.  Each
question record includes its video ID, question, options, answer, source
category, and annotated temporal evidence.  We uniformly name the latter
\texttt{time\_reference}, including annotations originally stored as
\texttt{clue\_intervals}.  The release does not redistribute video media,
signed URLs, local paths, execution traces, or experiment artifacts; videos are
resolved through their original CG-Bench IDs.  Split membership is by video ID,
and we verify that no video occurs in both splits.

The eight distributions span distinct but internally recurring forms of video
evidence.  \textbf{Interview} contains interviews and dialogue-centered
interactions, requiring localization of spoken content, speakers, and local
visual interactions.  \textbf{Product Presentation} contains product
demonstrations and sales-oriented presentations, where questions repeatedly
depend on product attributes, displayed text, object orientation, and
object-relative positions.  \textbf{Software Tutorial} consists of screen-based
software instruction, requiring interface-text reading, operation ordering, and
state-change tracking.

\textbf{Dramatic Narrative} contains TV-drama and narrative scenes, which
require resolving characters, events, and relations across cuts.  \textbf{Stage
Performance} covers performances such as magic and stage interactions, where
event phases, ordinal relations, and brief actions are often decisive.
\textbf{Gameplay} contains game recordings with dynamic scene state, player or
object actions, and interface evidence.  \textbf{Sports Broadcast} comprises
sporting events and broadcasts, emphasizing subject tracking across views,
actions, scores, and temporally localized events.  \textbf{Course Lecture}
contains instructional lectures with slide-based material, requiring joint use
of lecture content, on-screen text, concepts, and page--speech timing.

Grouping videos into coherent distributions avoids conflating unrelated failure
mechanisms and mirrors domain-specific deployment. For example, e-commerce
agents may emphasize speech retrieval, product localization, and text
perception, whereas course-video agents may emphasize slide parsing, speech
transcription, and knowledge localization.

These descriptions characterize repeated evidence-localization and perception
requirements within a distribution; they do not impose a hand-written
question-type rule on individual examples.  Their purpose is to make it
possible to aggregate individual failures into a distribution-level design
diagnosis, while the variation across the eight tasks tests whether an evolution method produces
different functional adaptations.

\begin{table}[t]
\centering
\scriptsize
\begin{tabular}{lrrrr}
\toprule
Distribution & \multicolumn{2}{c}{Evolution} & \multicolumn{2}{c}{Held-out} \\
 & Videos & Questions & Videos & Questions \\
\midrule
Interview & 5 & 41 & 14 & 135 \\
Product Presentation & 5 & 50 & 12 & 103 \\
Software Tutorial & 5 & 50 & 14 & 132 \\
Dramatic Narrative & 5 & 50 & 12 & 119 \\
Gameplay & 5 & 43 & 9 & 85 \\
Stage Performance & 5 & 42 & 11 & 103 \\
Sports Broadcast & 4 & 44 & 8 & 71 \\
Course Lecture & 4 & 36 & 12 & 119 \\
\midrule
Total & 38 & 356 & 92 & 867 \\
\bottomrule
\end{tabular}
\caption{Final VA-EvoBench split sizes.  Videos are disjoint across the
evolution and held-out splits.}
\label{tab:split-statistics}
\end{table}

\begin{figure}[t]
\centering
\includegraphics[width=\columnwidth]{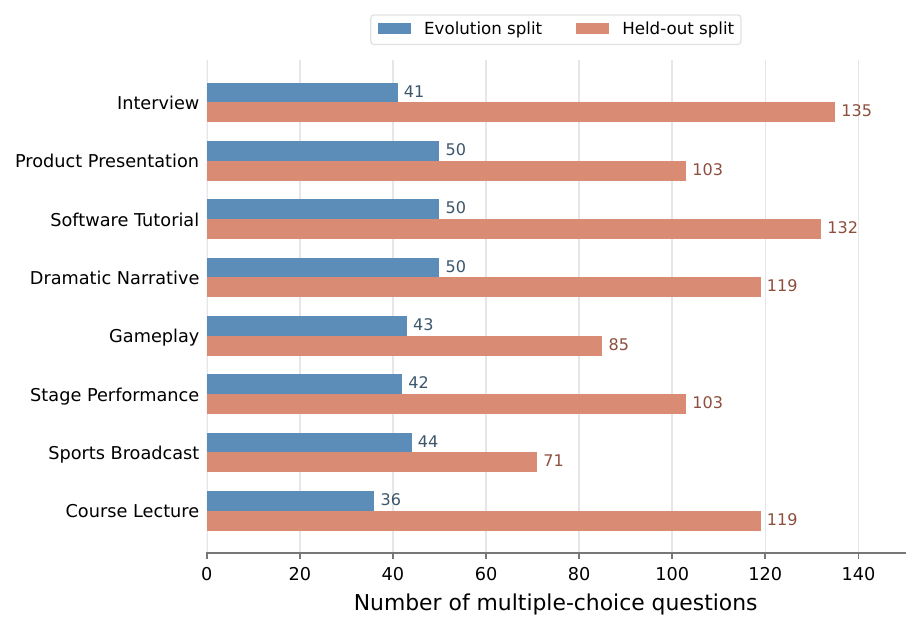}
\caption{Question counts for the eight independent evolution tasks. Every
target distribution has an evolution split used for adaptation and a disjoint
held-out split used exclusively for evaluation. Numbers are displayed beside
the corresponding bars.}
\label{fig:split_question_counts}
\end{figure}

The evolution splits contain 36--50 questions per type (356 in total), while
the disjoint held-out splits contain 71--135 questions per type (867 in total).
This allocation supplies recurring within-type failure evidence for adaptation
while retaining a larger independent set for outcome measurement.

\subsection{Mixed-Distribution Boundary Evaluation}
We additionally evaluate MVA on LVBench after evolving it on a mixed evolution
set sampled from the eight target distributions. The source videos used for
evolution are disjoint from all LVBench evaluation videos. This experiment
characterizes the boundary of distribution-adaptive evolution and is not part
of the primary VA-EvoBench setting.

\section{Additional Experimental Results}
\label{app:additional-results}

\subsection{Final Per-Distribution Outcomes}

Table~\ref{tab:per-distribution-outcomes} gives the exact final held-out
correct counts underlying the MVA row in Table~\ref{tab:main_results}.  The
last row is question-weighted; the main-paper Avg. remains the macro average.

\begin{table*}[t]
\centering
\small
\setlength{\tabcolsep}{4.0pt}
\begin{tabular}{lrrr}
\toprule
Distribution & Correct / total & Acc. (\%) & Initial $\rightarrow$ final (pp) \\
\midrule
Interview & 64 / 135 & 47.41 & $40.00 \rightarrow 47.41$ \\
Product Presentation & 58 / 103 & 56.31 & $48.54 \rightarrow 56.31$ \\
Software Tutorial & 70 / 132 & 53.03 & $42.42 \rightarrow 53.03$ \\
Dramatic Narrative & 48 / 119 & 40.34 & $37.82 \rightarrow 40.34$ \\
Stage Performance & 59 / 103 & 57.28 & $23.30 \rightarrow 57.28$ \\
Gameplay & 40 / 85 & 47.06 & $31.76 \rightarrow 47.06$ \\
Sports Broadcast & 36 / 71 & 50.70 & $40.85 \rightarrow 50.70$ \\
Course Lecture & 71 / 119 & 59.66 & $42.86 \rightarrow 59.66$ \\
\midrule
All held-out questions & 446 / 867 & 51.44 & $38.75 \rightarrow 51.44$ \\
\bottomrule
\end{tabular}
\caption{Final held-out outcomes of MVA by target distribution.}
\label{tab:per-distribution-outcomes}
\end{table*}

\subsection{Iteration-Wise Evolution Trajectories}
\label{app:iteration-results}

Table~\ref{tab:iteration-trajectories} contains the values used for the
evolution-dynamics figure. Every entry is independently recorded for an
archived checkpoint on the held-out split and is analyzed only after the
evolution budget is exhausted. It is not an input to candidate generation,
promotion, or checkpoint selection. The macro average rises from
38.44\% for the initial agents to 51.47\% at the end of evolution, while
individual trajectories remain heterogeneous.

\begin{table*}[t]
\centering
\small
\begin{tabular}{lrrrrr}
\toprule
Distribution & Initial & Iter. 1 & Iter. 2 & Iter. 3 & Iter. 4 \\
\midrule
Interview & 40.00 & 39.26 & 37.78 & 40.00 & 47.41 \\
Product Presentation & 48.54 & 47.57 & 52.43 & 51.46 & 56.31 \\
Software Tutorial & 42.42 & 43.18 & 46.97 & 44.70 & 53.03 \\
Dramatic Narrative & 37.82 & 39.50 & 38.66 & 46.22 & 40.34 \\
Gameplay & 31.76 & 38.82 & 29.41 & 27.06 & 47.06 \\
Stage Performance & 23.30 & 42.72 & 42.72 & 43.69 & 57.28 \\
Sports Broadcast & 40.85 & 42.25 & 45.07 & 45.07 & 50.70 \\
Course Lecture & 42.86 & 53.78 & 56.30 & 57.14 & 59.66 \\
\midrule
Macro average & 38.44 & 43.39 & 43.67 & 44.42 & 51.47 \\
\bottomrule
\end{tabular}
\caption{Read-only held-out accuracy (\%) by evolution iteration.}
\label{tab:iteration-trajectories}
\end{table*}

\subsection{Evolution-Split Trajectories}

\begin{figure}[t]
\centering
\includegraphics[width=\columnwidth]{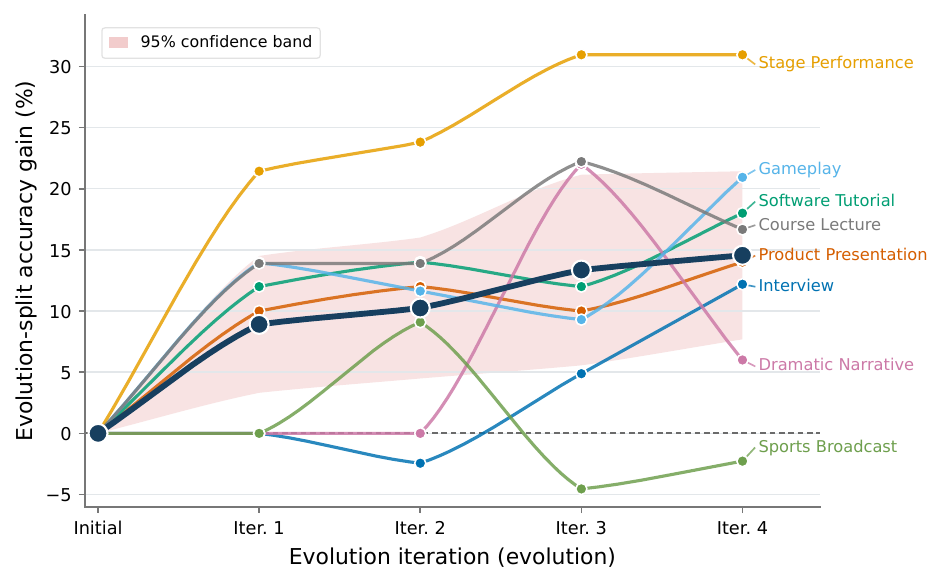}
\caption{Evolution-split accuracy trajectories across eight video
distributions.  The horizontal axis labels evolution checkpoints, whereas the
main-paper efficiency figure uses cumulative evolution compute.}
\label{fig:train_evolution_dynamics}
\end{figure}

Figure~\ref{fig:train_evolution_dynamics} reports the corresponding
evolution-split trajectories.  These values are included for process
transparency; held-out accuracy remains the outcome measurement reported in the
main table.

\subsection{Ablation Studies}
\label{app:ablation-trajectories}

All Course Lecture variants use the same 119-question held-out split and identical
question set.  Table~\ref{tab:ppt-ablation-trajectories} gives their full
read-only trajectories; every ablation result in the main paper uses the last
iteration of its corresponding trajectory.  The complete procedure improves from 42.86\% at
initialization to 59.66\% at the end of evolution. In contrast, blind evolution collapses to
22.69\% immediately after its first update before recovering unevenly.  This
volatility explains why its one relatively high intermediate score should not be read as
evidence that review-driven diagnosis and research are unnecessary.

Removing only Review is not a well-defined intervention: Diagnosis and optional
Deep Research consume review artifacts.  We therefore remove Review, Diagnosis,
and Deep Research jointly to create the blind-evolution condition.  Similarly,
a no-probe condition is deliberately not executed.  The minimal-validation
probe is both a fast test of a local design hypothesis and a runtime-observation
mechanism used to inspect the candidate's concrete trajectory; removing it
eliminates the defined repair-and-validation path rather than isolating an
ordinary performance component.

\begin{table*}[t]
\centering
\small
\begin{tabular}{lrrrrrr}
\toprule
Setting & Initial & Iter. 1 & Iter. 2 & Iter. 3 & Iter. 4 & Vol. \\
\midrule
Full MVA & 42.86 & 53.78 & 56.30 & 57.14 & 59.66 & 15.52 \\
w/o Diagnosis & 42.86 & 50.42 & 53.78 & 50.42 & 57.14 & 18.48 \\
w/o optional Deep Research & 42.86 & 39.50 & 47.90 & 46.22 & 48.74 & 20.59 \\
Blind evolution & 42.86 & 22.69 & 50.42 & 48.74 & 47.90 & 293.24 \\
MVA initialized from DVD & 60.50 & 43.70 & 54.62 & 52.10 & 63.03 & 131.43 \\
w/o minimal-validation probe & -- & -- & -- & -- & -- & -- \\
\bottomrule
\end{tabular}
\caption{Course Lecture ablation trajectories: held-out accuracy (\%).  All
measurements are read-only. Vol. is the population variance, in
\(\mathrm{pp}^2\), of signed consecutive changes
\(\delta_k=\mathrm{Acc}_k-\mathrm{Acc}_{k-1}\).}
\label{tab:ppt-ablation-trajectories}
\end{table*}

\begin{figure}[t]
\centering
\includegraphics[width=\columnwidth]{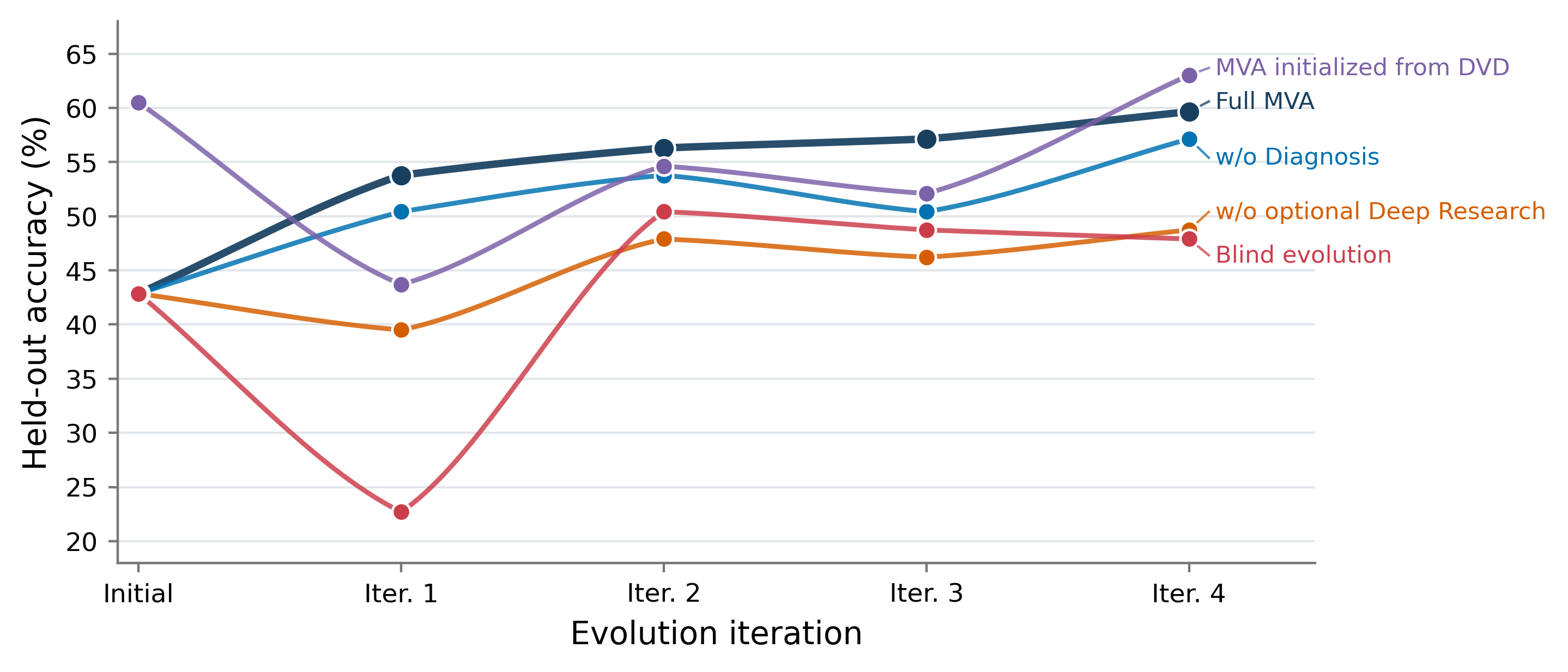}
\caption{Held-out evolution trajectories for Course Lecture ablations and
initialization settings. Results are used only for reporting.}
\label{fig:ablation-evolution-dynamics}
\end{figure}

\subsection{Evolution Efficiency and Cost Accounting}

The efficiency curve evaluates an evolution algorithm as a compute--accuracy
relation rather than prescribing a universal cost formula. In our
instantiation, the cumulative coordinate is the total model-token consumption
of the complete evolution process. Each run is normalized by its own final
cumulative cost before aggregation. At a checkpoint, the reported vertical
value is the best-so-far evolution-split accuracy among agents produced within
that budget. This curve is descriptive only and never changes the final-agent
protocol or selects a held-out checkpoint.

\section{Qualitative-Case Interpretation}
\label{app:qualitative}

We interpret the three cases in Figure~\ref{fig:qualitative_cases}, one
additional Dramatic Narrative case, and one failed Gameplay candidate using
their executed evidence trajectories.  Each successful comparison follows an
earlier failed combo and a later combo that applies a reusable behavior induced
from the corresponding evolution split.  These displayed held-out questions
serve only as post-hoc illustrations of the resulting distribution-level
behaviors.  The Gameplay example is instead an evolution-split regression
exposed by Review and rejected before candidate promotion.

\paragraph{Event ordering in Stage Performance.}
For the first-ball question, the earlier combo inspected 90--100\,s,
390--400\,s, and 560--570\,s.  These ranges contained broadly related magic
scenes, but the combo neither identified which performance event they belonged
to nor preserved their order, and it therefore could not resolve ``the first
ball in the first trick.''  The later combo conditionally constructs a
time-stamped global event directory when the planner determines that local
evidence is insufficient for an ordinal or relational query.  It then performs
field-aware event retrieval while an evidence ledger retains event hypotheses,
candidate ranges, inspected ranges, and observations.  In this execution, the
global browse covered the opening 0--99.375\,s and exposed the early
white-ball event, after which the agent considered localized event candidates
at 40--45\,s and 55--60\,s and answered \emph{White}.  The evolved behavior is
not a rule for this particular ball color: it is a control strategy for
disambiguating repeated stage events by their temporal order.

\paragraph{Relation-conditioned verification in Product Presentation.}
The earlier combo already retrieved the relevant jersey display at
1140--1150\,s, together with an irrelevant outro window at 2060--2070\,s, but
committed to the ambiguous reading ``ARCHIVE.''  The later combo keeps the
requested back-side text as an unresolved fact rather than immediately mapping
an uncertain reading to an option.  Localization preserves the strict product
anchors and may add a structurally linked local alternative for a named missing
fact; memory separately records seen ranges, acquired facts, and unresolved
facts; and a fresh range must be perceptually verified before the agent can
finish.  The resulting trajectory considered 10--20\,s, 310--320\,s,
1140--1150\,s, and 1150--1160\,s before selecting ``RCHIVES.''  Thus, the
reusable behavior is relation-conditioned evidence acquisition: the agent
first seeks the requested object view---here, the back of the jersey---and
only then resolves its fine-grained attribute.

\paragraph{Identity-anchored verification in Sports Broadcast.}
The earlier combo mixed observations from different players and broadcast
views.  It combined a black-kit observation with a later red-shirt/white-shorts
view of Lin, and promoted an unresolved wrist detail into an unsupported black
wristband claim, yielding the incorrect statements 2 and 4.  The later combo
uses the on-screen ``LIN'' identity as an anchor, associates overview
observations and local regions with that provenance, and carries unresolved
attributes across views.  It localized compact attire windows at 660--670\,s
and 740--750\,s.  The first overview proposed a subject region, and a tracked
crop over 668--670\,s was used to test the wristband hypothesis.  Because that
crop returned no usable wrist observation, the attribute remained unresolved
rather than becoming positive evidence.  A second ``LIN''-anchored view then
showed a pinkish-red short-sleeved shirt and predominantly white shorts with
red trim, without establishing a black wristband, leading to statements 1 and
2.  The contribution of the tracked crop is therefore evidence control, not a
claim that the crop itself recognized the wrist: together with the identity
anchor and cross-view memory, it prevents attributes of another athlete from
being assigned to Lin.

\paragraph{Constraint-aware recovery in Dramatic Narrative.}
For the question asking what the woman in red was doing while the man beside
her clapped, the earlier combo performed one retrieval--verification pass over
50--60\,s, 180--190\,s, and 190--200\,s.  None contained the target
interaction, yet the combo guessed that she was sitting on the ball and
clapping along.  The later combo instead represents the question as a
constraint-bearing evidence specification: it records the two target entities,
the man's clapping as the trigger action, the woman's concurrent action as the
missing fact, and the required visual and speech channels.  Its first candidate
windows, 235--240\,s and 430--435\,s, failed these joint constraints and were
recorded as excluded evidence.  A bounded recovery then produced
720--725\,s and 740--745\,s; the former shows the man clapping while the woman
in red sits on a pink exercise ball and drinks from a clear bottle.  The combo
therefore answers that she was sitting on the ball and drinking water.  This
behavior generalizes to narrative questions whose answer depends on verifying
that multiple characters, actions, and temporal relations co-occur in the same
scene, rather than merely retrieving windows containing similar keywords.

\paragraph{Over-specialized regression in Gameplay.}
One rejected candidate shows how an apparently useful specialization can
become an over-specialized hard classifier.  The question asks which
statements are correct when the protagonist first enters a space station.  It
requires three kinds of fine-grained visual evidence: whether the backpack
ring has two colors, which shoulder carries a spherical robot, and whether the
passage is square.  The current-best combo and the candidate retrieved the
same 70--80\,s, 110--120\,s, and 190--200\,s windows.  Although these ranges
missed the annotated 11--23\,s interval, the 70--80\,s window repeats the
entrance scene and contains the required attributes.  The current-best combo
passed the full question to scoped perception, which described a two-colored
ring, a round robot on the left shoulder, and a non-square octagonal passage,
supporting statements 1 and 2 (option E).

The evolved candidate instead classified any question containing
``which of the following'' or ``statements'' as an
\texttt{action\_statement} task.  Its perception prompt then restricted the
same visual windows to a fixed action ledger---\emph{sit}, \emph{play},
\emph{talk}, \emph{walk past}, or \emph{none observed}---without passing the
original attribute question.  Consequently, the observation discarded color,
relative-position, and shape evidence even though these details were visible.
Reasoning did not replan after receiving this mismatched evidence and selected
statements 3 and 4 (option A), turning a previously correct answer into an
error.  Review attributed the regression to the perception policy.  The
subsequent full evolution-split comparison exposed the broader regression, so
the acceptance gate rejected the candidate and preserved the prior current
best.  This case
illustrates why MVA evaluates generated specializations across the full
evolution split: a narrow policy that helps one surface form must not be
promoted when it reduces the evidence bandwidth required by neighboring
questions.

\section{Evolution Protocol and Annotation Access}
\label{app:evolution_protocol}

For a target type, MVA profiles only the evolution videos and their query set,
constructs an initial agent, and evolves it only on that type's evolution
split.  The executable Student observes only a video, a multiple-choice
question, and its answer options.  After an evolution-split execution, the
Teacher may use the source answer label and annotated evidence interval to
review an incorrect trajectory and construct diagnosis artifacts.  Source
annotations also retain the question, options, answer, evidence interval, and
available video and question metadata for reproducibility.

An external evaluator records held-out outcomes for archived checkpoints, but
these outcomes remain inaccessible to the evolution method throughout the
complete run. After the fixed $K$ iterations, we freeze and report the
last-iteration agent $A_K$, rather than selecting an intermediate checkpoint
according to held-out performance. Held-out outcomes are never used for review,
diagnosis, research, code generation, candidate promotion, early stopping, or
rollback.

\section{Detailed Evaluation Measures}
\label{app:evaluation_measures}

We report multiple-choice QA accuracy for each distribution, as well as tokens
per question (Tok./q) and image frames per question (Frm./q). The main-paper
Avg. is the macro average over the eight target distributions. Tok./q and
Frm./q follow the common per-question accounting implementation used for the
main comparison.

To evaluate evolution efficiency without tying the benchmark to a particular
agent-evolution design, we report cumulative compute consumed by the complete
evolution process. Let $c_j$ denote the model-token consumption at evolution
stage $j$. The cumulative evolution budget after stage $k$ is
\[
C_k = \sum_{j=0}^{k} c_j.
\]
In our experiments, the accounting unit is total model tokens. At each compute
checkpoint, we report the best accuracy achieved on the evolution split within
the available budget,
\[
P_d(C_k)=\max_{0\leq j\leq k}\mathrm{Acc}^{\mathrm{evol}}_d(A_j).
\]
Together, cumulative evolution compute, best-so-far evolution accuracy, and
per-question cost characterize performance obtained under a given evolution
budget. These evolution-split measurements do not alter the held-out protocol:
the main comparison always reports the final agent $A_K$.

\section{Discussion}
\label{app:discussion}

Multimodal intelligence has advanced rapidly~\cite{xu2026seeingnotthinking,wang2023openvocabularyobjectdetection,hu2021ramstrans,li2026perceivetoreason}.

The ablations clarify the role of the diagnostic stack. Cross-trajectory
diagnosis provides directions grounded in recurring failures, whereas blind
evolution is highly volatile. Minimal validation tasks test local changes and
exposes their execution traces before full evaluation, making it integral to
reliable evolution.

MetaVideoAgent targets relatively coherent distributions with recurrent
evidence requirements rather than arbitrary heterogeneous corpora. Its
effectiveness depends on review quality and module attribution, while full
long-video evaluation remains expensive. DVD initialization shows that
evolution can start from an existing agent, but does not establish
cross-distribution transfer. Broader mixtures may require latent-distribution
discovery and coordination of conflicting updates.

\section{Limitations}
\label{app:limitations}

MetaVideoAgent is designed for a target distribution with sufficiently
recurrent evidence requirements to support distribution-level diagnosis.  It
does not discover such distributions automatically, nor does it establish that
an agent evolved on one target distribution transfers to another.  In
particular, a broad mixture can combine incompatible failure modes and may
require latent-distribution discovery or explicit coordination of competing
updates before the present protocol applies.

Evolution can also settle on a locally effective but over-specialized update.
Diagnosis and the minimal-validation set deliberately concentrate on recurrent
failures, so a generated policy may encode a narrow task classifier, suppress
evidence needed by neighboring questions, or accumulate biases over later
iterations.  The Gameplay regression in Section~\ref{app:qualitative}
illustrates this failure mode.  Guard tasks and full evolution-split comparison
expose observed regressions, but a finite split cannot rule out distributional
overfitting.  Broader candidate comparisons and diversity-aware regularization
remain important safeguards for longer evolution horizons.

The quality of evolution is further bounded by the fidelity of review and
module attribution.  Incomplete review, ambiguous temporal references, or an
incorrect causal attribution can yield an unhelpful diagnosis contract.
Static checks, smoke execution, and minimal-validation probes reduce
engineering and local-runtime failures, but they cannot establish that a
proposed mechanism is the unique cause of an error.

Finally, evolution remains compute intensive because a candidate must execute
within a long-video agent before it can be assessed.  The reported four-update
setting is an experimental budget, not a convergence guarantee or a
framework-level limit on further evolution.  Extending the method to longer
evolution horizons, broader mixtures, and cross-distribution transfer requires
more efficient evaluation and stronger safeguards against accumulated
diagnostic error.

\clearpage

\clearpage

\end{document}